# Two-Stage Human Verification using HandCAPTCHA and Anti-Spoofed Finger Biometrics with Feature Selection

***Asish Bera [a]*, Debotosh Bhattacharjee [b,c], and Hubert P. H. Shum [d]***

*[a]Department of Computer Science, Edge Hill University, St. Helens Road, Ormskirk, Lancashire, L39 4QP, United Kingdom.*
*email: asish.bera@gmail.com / beraa@edgehill.ac.uk*

*[b]Department of Computer Science and Engineering, Jadavpur University, 188, Raja S.C. Mallick Rd, West Bengal 700032, India.*
*[c]Center for Basic and Applied Science, Faculty of informatics and management, University of Hradec Kralove, Rokitanskeho 62, 500 03 Hradec Kralove, Czech Republic.*
*email: debotosh@ieee.org*

*[d]Department of Computer Science, Durham University, Stockton Road, Durham, DH1 3LE, United Kingdom.*
*email: hubert.shum@durham.ac.uk*

ABSTRACT

This paper presents a human verification scheme in two independent stages to overcome the vulnerabilities of attacks and to enhance security. At the first stage, a hand image-based CAPTCHA (HandCAPTCHA) is tested to avert automated bot-attacks on the subsequent biometric stage. In the next stage, finger biometric verification of a legitimate user is performed with presentation attack detection (PAD) using the real hand images of the person who has passed a random HandCAPTCHA challenge. The electronic screen-based PAD is tested using image quality metrics. After this spoofing detection, geometric features are extracted from the four fingers (excluding the thumb) of real users. A modified forward-backward (M-FoBa) algorithm is devised to select relevant features for biometric authentication. The experiments are performed on the Boğaziçi University (BU) and the IIT-Delhi (IITD) hand databases using the *k*-nearest neighbor and random forest classifiers. The average accuracy of the correct HandCAPTCHA solution is 98.5%, and the false accept rate of a bot is 1.23%. The PAD is tested on 255 subjects of BU, and the best average error is 0%. The finger biometric identification accuracy of 98% and an equal error rate (EER) of 6.5% have been achieved for 500 subjects of the BU. For 200 subjects of the IITD, 99.5% identification accuracy, and 5.18% EER are obtained.

## 1. Introduction

Human interaction over social networks is proliferating nowadays to a more considerable extent (Madisetty and Desarkar, 2018). Individual authentication for social networking requires immense attention to thwart automated bot attacks such as spam detection in Twitter (Sedhai and Sun, 2017) and other social networks. Popular social networking sites (e.g., Facebook, etc.) use a *Completely Automated Public Turing Test to Tell Computers and Humans Apart* (CAPTCHA) to avert automated bots (Torky and Ibrahim, 2016). Unauthorized malicious bots (e.g., Mirai, Zbot, etc.) may create a fraudulent individuality for identity theft through these networking services. As a solution, an intelligent biometric system plays a vital role in enhancing security measures of social interaction (Sultana et al., 2017). Biometrics is inevitably used for secure individualization in real-time and internet-based services. Simultaneously, the vulnerabilities of various access control systems are growing extensively. The Internet Security Threat Reports (ISTR) on different intelligent threats, mentioned in Table-1, reflect that security aspects must be enhanced globally. The CAPTCHA tackles such threats by discriminating between human and automated programs (Cheng et al., 2019; Gao et al., 2019; Belk et al., 2015; Zi et al., 2020). A robust CAPTCHA design algorithm (CDA) can endow improved security, which is intractable to bots. The difference between the recognition ability of humans and machines defines the degree of confidence regarding security, i.e., the recognition gap between the human and machine. A CAPTCHA should be amply arbitrary and distorted so that bots cannot undermine it effortlessly (Datta et al., 2009), while it should be futile on human perception.

**Table 1: Some related information about ISTR, collected and provided by Symantec, offers an essential network for global cybersecurity data collection. The data mentioned in ISTR (vol. 22, 2017; vol. 23, 2018; vol. 24, 2019) are based on the U.S. and other countries.**

| Information type | Year-wise report |
|---|---|
| a) email found to be spam (%) | 53.4 (2016), 54.6 (2017), and 55(2018) |
| b) Data breaches due to phishing, spoofing, or social engineering | 21.8% (2015) and 15.8% (2016). The U.S. was affected mostly in 2016. Phishing level increased from 1 in 2995 emails (2017), to 1 in 3207 emails (2018) |
| c) Data breaches (identities stolen) | More than 7.1 billion online identities have been exposed in the last eight years |
| d) Attacks against IoT devices | Total 57691 (2017), and 57553(2018) |

Among several physiological traits, hand geometry is deployed for automatic identity verification based on different types of geometric and shape-based features (Dutağaci et al., 2008; Bera et al., 2014). The hand anatomy indicates that common geometric features are the lengths and widths of fingers, palm features, etc. Alternatively, the moment invariants (Baena et al., 2013), Fourier descriptors (FDs) (Kang and Wu, 2014), wavelet transform (Sharma et al., 2015), and scale-invariant feature transform (SIFT) (Charfi et al., 2014) represent shape-based features. Geometrical attributes are more easily to measure than silhouette based features. A combination of geometric and shape-based features can render good accuracy (Baena et al., 2013). Recently, deep learning-based features are computed from hand images in (Afifi, 2019). Besides feature extraction, relevant feature selection is pertinent to optimize feature space and improve the accuracy (Bera and Bhattacharjee, 2020; Kumar and Zhang, 2005; Reillo et al., 2000).

Nevertheless, the human face, fingerprint, iris, hand, and others are jeopardized by presentation (a.k.a. spoofing) attacks through printed photos, electronic screen display, etc. (Farmanbar and Toygar, 2017; Korshunov and Marcel, 2017; Nogueira et al., 2016; Tolosana et al., 2019). The hardware (at sensor module) and software (bypassing feature extraction, and other modules of a biometric system) based attacks (e.g., hill climbing attack) on biometrics have been addressed in (Uludag and Jain, 2004). Corresponding counter-measurements are explored using modern sensor technologies (hardware-based) and additional information processing (software-based) strategies. The image quality metrics ($IQ_m$) are beneficial for presentation attack detection (PAD) (Galbally et al., 2014).

It is averred that CAPTCHA and biometrics both are independently vulnerable to attack. Though biometrics can recognize a lawful person, it cannot distinguish whether a claimer is a human or a bot. Also, biometrics is susceptible to a spoofing attack for which a PAD method is essential for designing a secure CDA (Uzun et al., 2018). As a possible solution, we design a two-stage human verification process that judiciously conflates the computational intelligence of CAPTCHA and biometrics to thwart different attacks. It should be noted that spoofing detection and finger biometric verification (FBV) are considered here as a single stage for simplicity. The sequence of two stages, i.e., CAPTCHA, is followed by biometrics with PAD, which is significant because the earliest aim is to hinder the bots and ensure that any bot does not forge biometrics. These two different tools work together complementarily to alleviate intrinsic limitations to determine the bot and intruder. A PAD method detects and allows the real samples of the trait for authentication. The independent nature of both stages may not alter overall accuracy even if the stages are interchanged. However, the challenges of detecting a bot for biometric verification remain open if the biometrics is tested first. This proposed two-fold system is conceptualized in Fig.1.

In our earlier conference paper (Bera et al., 2018), a simple two-stage human verification process has been proposed that leverages both the benefits of CAPTCHA and biometrics to enhance the security, and it showcases some elementary results. In this proposed work, a novel method for spoofing attack detection using quality metrics is newly introduced at Level-2 (which was not described in (Bera et al., 2018)) to strengthen the security of the system by allowing only legitimate users. We propose a new feature selection method that aims at optimizing the number of features required and minimizing the computation time for online-based verification. Such an algorithm optimizes feature subset cardinality by selecting salient global and local features of four fingers to achieve better performances on two publicly available databases, namely, the BU (Yörük et al., 2006) and IIT-Delhi (IITD) (Kumar, 2008). The overall verification process requires a shorter computation time (<16s), which is essential for online verification. We have conducted extensive new experiments such as image quality assessment using $IQ_m$, permutation importance of each selected feature using random forest (RF) classifier, human verification with a disjoint population, etc. The HandCAPTCHA algorithm at Level-1demonstrates the security benefits over several attack schemes, which is a crucial objective of a CDA. In-depth analyses of algorithmic superiority to futile machine recognizability compared to the existing approaches regarding the false accept rate (FAR), probability estimation of attack, computation time, design parameters, and others are remarkable. Level-1 is presented with a new mathematical explanation to exhibit the strength of the proposed algorithm. Overall, comprehensive experimental results at each stage demonstrate the novelty of the proposed approach. The motivation of developing such a two-fold verification method within a shorter time-period to strengthen the security of various applications (such as internet of things (IoT), antispoofing, etc.) and utilizing both CAPTCHA and biometric modules complementarily is innovative in this research direction.

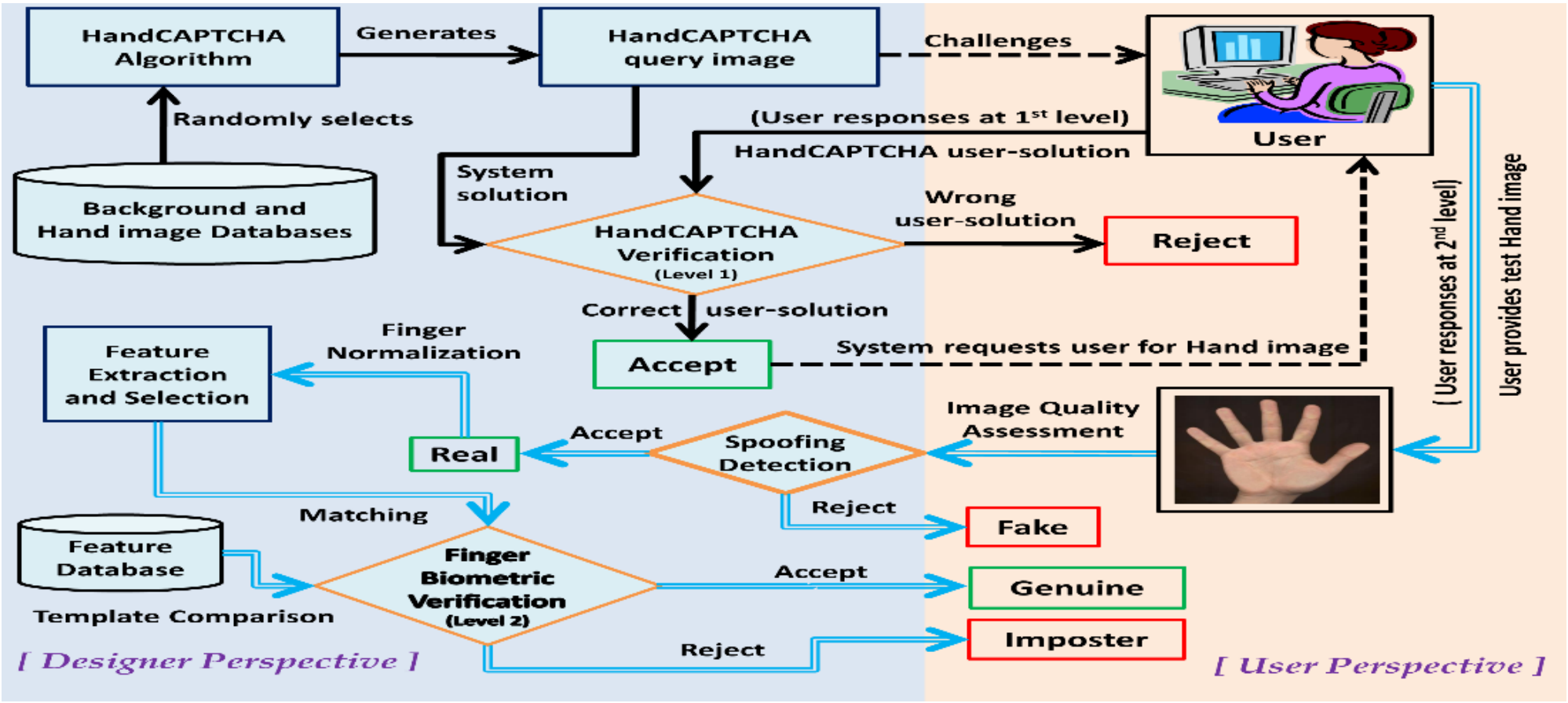

**Fig.1.** The proposed two-stage verification scheme. Level-1 performs HandCAPTCHA verification. Level-2 performs finger biometric recognition with spoofing attack detection. The left side represents the designer's aspects, and the right side represents the user's aspects. The real hands of the BU dataset are used for $I_{HCAPTCHA}$ generation, anti-spoofing, and biometric authentication. The two-levels are computationally independent regarding the preprocessing of various images.

As a summary, the proposed HCA generates an image-based CAPTCHA, namely, $I_{HCAPTCHA}$. Level-1 determines whether a human or bot solves an $I_{HCAPTCHA}$. If the responder is a human, then Level-2 recognizes the real hand of a person who claims to be a valid identity using geometric features of the four fingers, except the thumb. Level-1 is presented with a new experimental setup emphasizing usability study with wider variations in design parameters, and the significant security aspects are analyzed rigorously. Level-2 presents a PAD scheme using $IQ_m$ and then a modified forward-backward (M-FoBa) feature selection method for FBV.

The left- and right-hand images of the BU dataset are considered as the real hands in our gallery for HCA. For biometric verification, the left-hand images of BU and IITD are tested. Alternatively, the right-hand images or both types of hands images be used for experiments. A random subset of the BU dataset is used to generate fake hand artifacts for electronic screen-based spoofing-attack simulation, as followed in (Raghavendra and Busch, 2015). The dual purpose of reusing the same hand images of BU in both levels is not mandatory, and other datasets can be used at each level. It means that the datasets can be utilized independently for each verification task. A person who solves an $I_{HCAPTCHA}$ correctly is permitted for PAD and FBV at Level-2. This model offers two-levels of stringent security by circumventing unauthorized access through automated attacks compared to a single level. The main contributions of this paper are:

a) An electronic screen display based spoofing attack detection method using image quality metrics is presented. This anti-spoofing method allows only real hand images for finger biometric authentication by outwitting fake images. For experiments, a fake hand dataset is created using the original hand images of the BU dataset.
b) We propose a new feature selection method that aims at optimizing the number of features required and minimizing the computation time for online-based verification. The M-FoBa algorithm selects a combination of discriminative features from four fingers, except the thumb. Comprehensive experiments are conducted on two publicly available BU and IITD hand datasets.
c) The HandCAPTCHA algorithm demonstrates the security benefits over several attack schemes, which is our primary objective. In-depth analyses of algorithmic superiority to futile automated attacks regarding the false accept rate, probability estimation of attack, computation time, and others are described rigorously.

d) The utility of fundamentally two different recognition methods is conflated systematically, enhancing the security level and robustness of the system within a shorter period. Overall, this new model can restrain a bot and permit a human claimant whose actual finger geometry determines whether he/she is a genuine or fraud.

The rest of this paper is organized as follows. Section 2 describes related works on image-recognition CAPTCHA (IRC) and hand biometrics. Section 3 presents CDA and analysis of various automated attacks. The assessment of image qualities with $IQ_m$ is addressed in Section 4. Finger geometry with feature selection is briefed in Section 5. The experiments for verification stages are discussed in Section 6. The conclusion is drawn in Section 7. A list of frequently used abbreviations and symbols are defined in Table 2.

## 2. Related Work

### *2.1 Study on Image-based CAPTCHA*

Various CDAs have been formulated based on text, image, audio, video, cognitive, etc. The IRC is independent of native language (reading/writing), and also a suitable alternative than text-recognition CAPTCHA due to its robustness (Tang et al., 2018). A study on IRC is briefed in Table 3, and the numbers of images in respective CDAs are given within braces in col 3.

**Table 2: List of frequently used abbreviations and symbols**

| Name | Abbreviation | Name | Symbol |
|---|---|---|---|
| Average error rate | AER | Candidate distortion of type k (size/rotation/translation) | $\delta_k$ ($\delta_{size}/\delta_{rotation}/\delta_{translation}$) |
| Completely Automated Public Turing Test to Tell Computers and Humans Apart | CAPTCHA | Image of type k (real/fake/spoofed/background) | $I_k$ ($I_G$ / $I_F$ / $I_{SP}$) |
| | | An image database of type k (real/fake/background) | $D_k$ ($D_R$ / $D_F$/ $D_B$) |
| CAPTCHA design algorithm | CDA | HandCAPTCHA image | $I_{HCAPTCHA}$ |
| Equal error rate | EER | Entropy | $H(*)$ |
| False accept rate | FAR | Feature of type k (global/local) | $f_k$ ($f_g$ / $f_l$) |
| False reject rate | FRR | Probability of human and machine recognizability | $p\mathcal{H}(Q)$ and $p\mathcal{M}(Q)$ |
| Finger biometric verification | FBV | *out-of-bag* error and cross-validation error | $E_{OOB}$ and $E_{CV}$ |
| Forward-Backward | FoBa | Dimension of $I_{HCAPTCHA}$ | $S_{HC}$ |
| Hand image-based CAPTCHA | HandCAPTCHA | *Pearson* correlation coefficient | $\eta$ |
| HandCAPTCHA algorithm | HCA | Set of {*Index, Middle, Ring, Little*} fingers | $\psi$ |
| Independent component analysis | ICA | The input feature set and optimal feature subset | F and $F_{0pt}$ |
| Image quality metric | $IQ_m$ | Probability of an event k (total probability) | $p_k$ ($p_T$) |
| Image-recognition CAPTCHA | IRC | Object segmentation complexity | $\mathbf{O}(\acute{S})$ |
| Modified forward-backward | M-FoBa | Standard deviation | $\sigma$ |
| Out-of-bag error | OOB | Recognisability gap | $\Delta$ |
| Presentation attack detection | PAD | Shape-objects of type k (circle/rectangle/star) | $\tau_k$ ($\tau_C/\tau_R/\tau_S$) |
| Receiver operating characteristic | ROC | Permutation importance | $\mathcal{P}$ |

**Table 3: A Study on State-of-the-Art Image Recognition CAPTCHAs, Presenting about Usability and Security Analysis**

| Ref. | CAPTCHA Description | Design Specification (#no of imgs.) | Accuracy (%) | Time (s) | Random Guess Probability (%) |
|---|---|---|---|---|---|
| Torky et al., 2016 | Necklace CAPTCHA: a graph of *n* vertices (order *n*) with cyclic left & right shift rotation | Necklace puzzle (*n*=3 to 6) produces an n-bit string of number, alphabet, and symbol | 97.26 | 24 | 22.56-31.25 |
| Bera et al., 2018 | HandCAPTCHA: select two real hands from a complex background | Distortions are applied on real (2) and fake (5-7) hands | 98.34 | × | 0.47- 1.5 |
| Tang et al., 2018 | Style Area CAPTCHA: neural style transfer technique. Relies on semantic information understanding and pixel-level segmentation | Synthetic and style transferred regions (4-7) of random shapes (rectangle, leaf, etc.) | 93.1 | 9.73 | $4.88\times10^{-3}$ |
| Osadchy et al., 2017 | Deep Learning CAPTCHA: creating and adding immutable adversarial noise on the image | Images (12) of natural objects, shapes, textures, etc. | 86.67 | 7.66 | 0.7 |
| Gao et al., 2016 | Annulus CAPTCHA: count the number of the annulus in a complex background | Distorted geometric annulus (1-8) | 89 | 8.89 | $3.05\times10^{-3}$ |
| Conti et al., 2015 | CAPTCHaStar: recognize space by moving the cursor inside a drawable space containing stars | Small white squares (star), placed in a squared black space to form a shape | 90.2 | ≈27 | 0.09 |
| Datta et al., 2005 | Imagination: two-round click-and-annotations | A set of 8 random objects with annotation | 95 | × | $6.2\times10^{-5}$ |
| Goswami et al., 2014 | Face detection CAPTCHA: mark human faces | Distorted human faces (2-4) and fake faces | 98.5 | × | 0.237 |

The human faces are used in faceCAPTCHA (Goswami et al., 2012) and faceDCAPTCHA (Goswami et al., 2014). Though, the face as a biometric trait and security analyses are not explored in those works. The robustness of faceCAPTCHA is presented in (Gao et al., 2015). Plausible attacks using face detection and recognition are tested. It is reported that rotation and blending have a significant impact on extracting the face region, and hence, time-consuming. An average 48% of success has been achieved for attacking faceDCAPTCHA within 6.2s. Also, deep learning tools play a significant role in the robustness study of CDA (Tang et al., 2018). Deep learning-based IRC, namely, DeepCAPTCHA, is robust against various attacks with a tolerable limit of 1.5% FAR (Osadchy et al., 2017). A real-time CDA is presented to defend spoofing attacks (Uzun et al., 2018). Necklace CAPTCHA is another IRC for securing social networks (Torky et al., 2016). Multibiometric traits (face, fingerprint, and eyes) are useful to improve the security of CDA (Powell et al., 2016). An IRC based on hand images, namely, HCA, is presented in this work.

### *2.2 Study on Spoofing Detection using Hand Biometrics*

Multimodal biometric traits are useful measures against attacks (Uludag and Jain, 2004). The spoofing attacks using the face, iris, and fingerprint are tested (Chingovska et al., 2014). The IQM has been tested using the face, iris, and fingerprint samples for PAD in (Galbally et al., 2014). Also, IQMs are computed for spoofing detection using the face and palmprint images in (Farmanbar and Toygar, 2017). Spoofing attacks on hand geometry have been briefed with two different methods (Chen et al., 2005). First, a fake plaster hand of a valid user is made, presented to the sensor of a deployed system. Second, the handshape of a legitimate user is printed out in a hard copy (i.e., print attack) and presented to the sensor for verification. Bartuzi and Trokielewicz (2018) have proposed a spoofing attack detection method using the thermal and color hand images. Inspired with these methods, we have explored PAD using IQM on hand biometrics in this paper.

### *2.3 Study on Hand Biometrics with Feature Selection*

Feature selection is crucial for the performance enhancement of a biometric system. A summary of hand biometric approaches with feature selection is presented in Table 4. Also, the methods tested on the BU database are studied as the same dataset is tested in this proposal. A verification system using the SIFT features is described in (Charfi et al., 2014). The SIFT is attractive due to its invariance properties of rotation, scaling, and others. Hand-shape matching using the modified Hausdorff distance and independent component analysis (ICA) is described in (Yörük et al., 2006). The ICA is applied to extract shape information from normalized binary hand contour. Based on ICA, a similar work is presented in (El-Sallam et al., 2011). Also, hand biometric methods developed at various levels of fusion are studied (Bera et al., 2015). A score-level fusion with four fingers is developed using cumulative angular function based FDs, computed from finger contour and finger area (Kang and Wu, 2014). Feature selection using the genetic algorithm on the IITD dataset is described in (Baena et al., 2013). Recently, four finger-based authentications using forward-backward feature selection on the BU dataset is presented (Bera and Bhattacharjee, 2020). It is inferred that geometric features are attractive due to small template size and less computational time, essential for online verification. Thus, we have employed finger geometric features in this work regarding computational time.

**Table 4: A Study on the Hand (or Finger) Biometrics with Feature Selection Approaches**

| Ref. | Feature definition and selection | Feature selection approach | Experimentation: database and results |
|---|---|---|---|
| Baena et al., 2013 | 403 geometric and shape-based features (lengths, widths, rectangularity, moment invariants, FDs, etc.) are defined. No. of selected features: about 50 | A genetic algorithm is applied for feature selection, and mutual information determines the correlation between a pair of features, and to eliminate redundancy among features. | Databases: GPDS (144 subj.), IITD (137 subj.), and CASIA (100 subj.). Identification: about 97% with the latter two databases using GA-LDA. EER: 4 to 5%. |
| Bera and Bhattacharjee, 2020 | 30 geometric features per finger (area, solidity, extent, widths, distances from the centroid, etc.) are computed. No. of selected features per finger: 12 | The rank-based forward-backward method determines feature relevance. Firstly, forward selection, and then backward elimination is followed. | Database: BU (638 subjects). Identification: 96.56% with the four fingers of the right hand, using the random forest classifier. Equal Error Rate (EER): 7.8%. |
| Kumar and Zhang, 2005 | Twenty-three geometric features (lengths & widths of palm, and fingers, perimeter, etc.) are extracted. No. of selected features: 15 | *Pearson* correlation reduces the number of features with maintaining similar results. | Database: 100 subjects. Identification: 87.8% using the logistic model tree classifier. |
| Reillo et al., 2000 | Thirty-one geometric features (heights and widths of fingers, inter-finger angles, etc.). No. of selected features: 25 | The class variability ratio determines the feature relevance using the inter-class variability and intra-class variability. | Database: 20 persons. Classification: 97%, and approx. 10% EER using the Gaussian mixture model (GMM). |

## 3. HandCAPTCHA Design and Security Analysis

The HCA design parameters are irreproducible and highly random, i.e., once these parametric values are generated, it is harder to replicate the algorithmic operations with the same values. So, each $I_{HCAPTCHA}$ is independent and dynamic. The random design parameters reduce $p_{\mathfrak{M}}(Q)$. The security facets are analyzed in depth against the adversary attacks. The HCA follows successive distortions using a random combination of a background image ($I_B$), two real ($I_G$), and the utmost seven fake ($I_F$) hand images to create an $I_{HCAPTCHA}$.

$$I_{HCAPTCHA} \leftarrow \text{HandCAPTCHA}\left(\mathfrak{R}\{I_B \cup \{I_G^i\}_{i=1}^{2} \cup \{I_F^j\}_{j=1}^{7}\}\right) \quad (1)$$

Let Q denote the correct positions of two real hand image ($I_G$) out of nine unique locations $Q=\{q_1, q_2 \in \mathbb{N} \mid 1 \le \mathbb{N} \le 9\}$. Permutation of ${}^{9}P_{2}$ provides 72 possibilities that Q may choose one at a time. The human recognizability to identify Q is denoted by $\mathcal{H}(Q)$. The solution ($\mathbb{S}$) of an $I_{HCAPTCHA}$ is defined as

$$\mathbb{S}(I_{HCAPTCHA}, Q) = \begin{cases} Correct\,(accept) & if \quad \mathcal{H}(Q) = Q \\ Wrong\,(reject) & otherwise \end{cases} \quad (2)$$

Let $p_{\mathcal{H}}(Q)$ and $p_{\mathfrak{M}}(Q)$ signify the probabilities of a correct $I_{HCAPTCHA}$ solution by a human $\mathcal{H}$, and an automated algorithm (machine) $\mathfrak{M}$, respectively. The recognizability gap ($\Delta$) is

$$\Delta = p_{\mathcal{H}}(Q) - p_{\mathfrak{M}}(Q) > 0. \quad (3)$$

The objective is to maximize $\Delta$ by maintaining $p_{\mathcal{H}}(Q)$ and reducing $p_{\mathfrak{M}}(Q)$ using HCA with randomized operations to evade machine recognition probability.

$$\underset{p_{\mathcal{H}}(Q) > p_{\mathfrak{M}}(Q)}{maximize \ \Delta} \ \text{s. t.} \ random\,(HandCAPTCHA) \rightarrow I_{HCAPTCHA} \quad (4)$$

The randomness of a variable is measured regarding the entropy H of a discrete random variable V, is given by

$$H(V) = -\sum_{v \in V} p(v).\log p(v) \quad (5)$$

The conditional entropy of a pair of the discrete random variable (U, V) is defined as

$$H(U|V) = -\sum_{v \in V} \sum_{u \in U} p(u,v).\log p(u|v) \quad (6)$$

For maintaining $p_{\mathcal{H}}(Q)$, the randomness (entropy) of intra-class $I_G$ should be small. Contrarily, the entropy of inter-class $I_F$ should be large. As human perception is more resilient to distortion. Therefore, composite distortion with adversarial noise is applied to obfuscate the vision of the machine than human.

**Algorithm 1:** HandCAPTCHA generation

***Input:*** Number of genuine hand images ($n_G$), the number of fake hand images ($n_F$), number of shapes ($N_{shape}$); Databases: $D_B$, $D_G$, and $D_F$
***Output:*** HandCAPTCHA image $I_{HCAPTCHA}$

$\ell_{I_B} \leftarrow \ell(D_B)$
$L_{Shape} \leftarrow \{ \tau_C(N_{Shape}(x_i, y_j)), \tau_R(N_{Shape}(x_i, y_j)), \tau_S(N_{Shape}(x_i, y_j)) \mid (N_{Shape}(x_i, y_j)) \in S_{HC}\}$
$\acute{L}_{Shape} \leftarrow \{\tau_{RGB}(L_{Shape}), \tau_{opacity}(L_{Shape})\}$
$\bar{I}_B \leftarrow \delta_{noise}(\delta_{morph}(\delta_{overlay}(I_B, \acute{L}_{Shape})))$
$I_{BG} \leftarrow \bar{I}_B$
$t_{I_G} \leftarrow \ell(D_G)$
*for i=1 to $n_G$*
  $\bar{I}_G^{(i)} \leftarrow \delta_{translation}\left(\delta_{rotation}\left(\delta_{size}\left(I_{t,G}^{i}\right)\right)\right)$
  $\bar{I}_B \leftarrow \delta_{overlay}\left(\bar{I}_B, \bar{I}_G^{(i)}\right)$
*end*
*for j=1 to $n_F$*
  $u_{I_F} \leftarrow \ell(D_F)$
  $\bar{I}_F^{(j)} \leftarrow \delta_{translation}\left(\delta_{rotation}\left(\delta_{size}\left(I_{u,F}^{j}\right)\right)\right)$
  $\bar{I}_B \leftarrow \delta_{overlay}\left(\bar{I}_B, \bar{I}_F^{(j)}\right)$
*end*
$I_{HCAPTCHA} \leftarrow \delta_{\alpha\text{-blend}}(\bar{I}_B, I_{BG})$
$I_{HCAPTCHA} \leftarrow \delta_{\gamma\text{-correction}}(I_{HCAPTCHA})$
*return $I_{HCAPTCHA}$*

***Major cues to solve $I_{HCAPTCHA}$ correctly at the user level***
a) Hand size and shape. b) Hand gadgets (like a ring, wristwatch, bracelets, sleeves, etc.). c) Special identification marks like tattoos. d) Major palm lines.

e) Inter-finger spacing status. f) Skin color. g) Male/Female. h) Similar background. i) Dissimilarities with the fake hand images by human perception.

### 3.1 HandCAPTCHA Algorithm (HCA) Design

Three different image sets, namely, the background ($D_B$), genuine hand ($D_G$), and fake hand ($D_F$) databases, are used in HCA. First, a background image ($I_B$) is processed with random distortions. Then, consecutive steps are followed to process $I_G$ and $I_F$, overlaid on the background at random spatial locations (Fig.2). The *step* number of *Algorithm-1* is mentioned accordingly for easier readability. In HCA (*Algorithm-1*), a composite distortion is applied, and three more $I_F$ are included to improve robustness against guessing attack compared to (Goswami et al., 2014; Goswami et al., 2012). The α-blending and γ-correction are varied to enhance the intensity and generate a challenging $I_{HCAPTCHA}$, shown in Fig.3.

#### 3.1.1 Background Image Preprocessing

The dimension ($S_{HC}$) of $I_{HCAPTCHA}$ regarding width ($X$) and height ($Y$) is 460×460 pixels. It is permissible to configure at a scalable extent according to a deployable device. The dimension of $I_B$ is also 460×460 pixels. $N_B$ denotes the cardinality of the background image set. Every $I_B$ is assigned to an unambiguous label. A random label-index ($\ell$) is generated to select a labeled background image $I_B$ from $D_B$ (*step* 1 of *Algorithm-1*).

$$\ell_{I_B} \longleftarrow \ell(D_B) \quad \text{where } \ell \in N_B \tag{7}$$

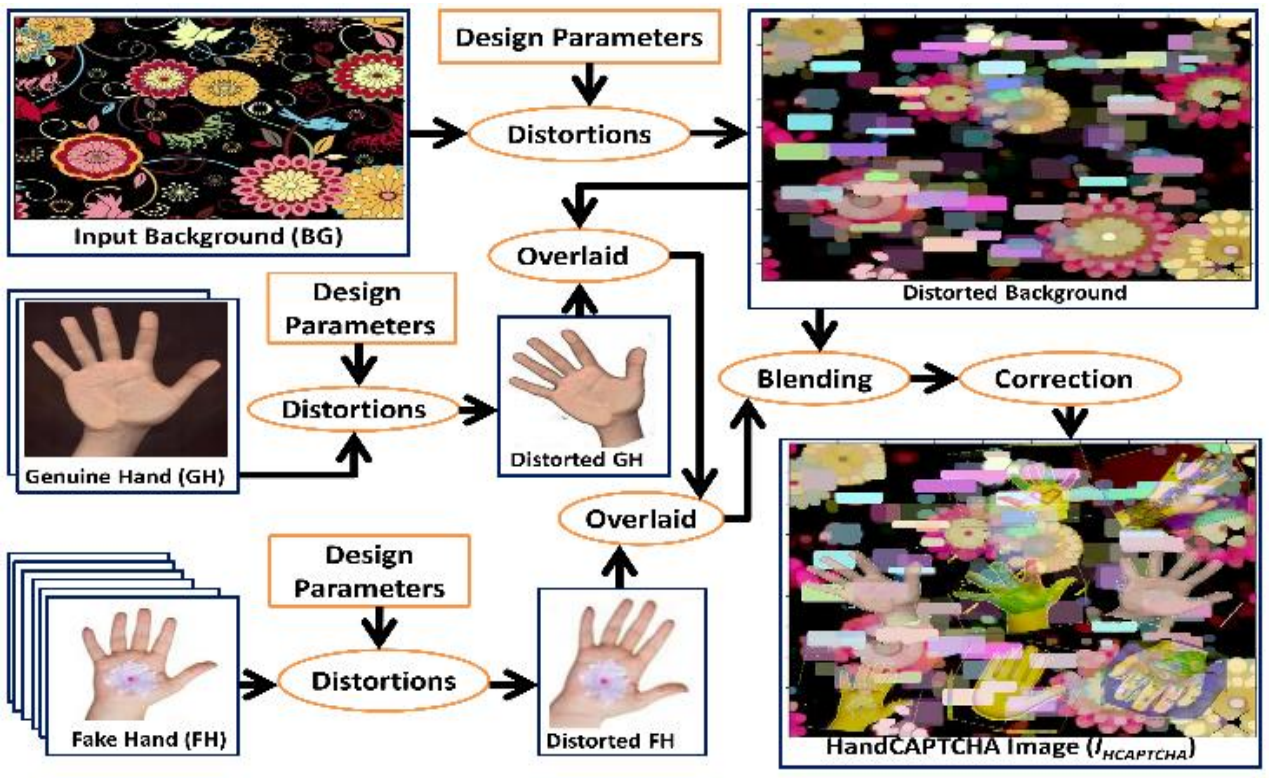

**Fig.2.** Key steps of the HandCAPTCHA algorithm.

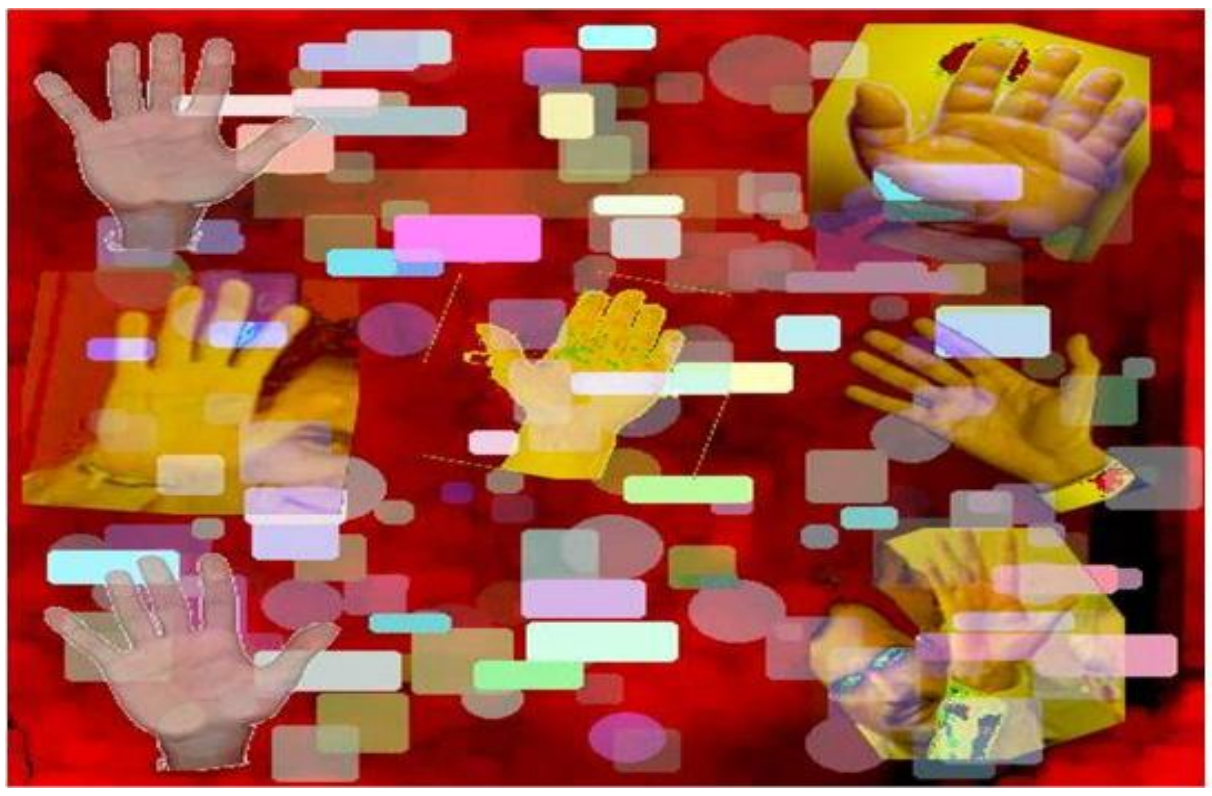

**Fig.3.** Sample $I_{HCAPTCHA}$ image

Three subsets of the total 300 random shapes with the initial pixel positions ($x_i$,$y_j$) are selected in *step* 2. The $N_{shape}$ numbers of different shape-objects, mainly, circle ($\tau_C$), rectangle ($\tau_R$), and star ($\tau_S$) with variable dimensions up to a maximum of 30-pixels are chosen arbitrarily. For each shape, 100 objects are considered. These $N_{shape}(x_i,y_j)$ positions should lie within $S_{HC}$. These positions are unique for any particular form (intra-shape) and almost unique in different ways (inter-shape). Though, the least numbers of pixel overlapping (i.e., one shape may be partially superimposed on another shape) between two types may be tolerable to avoid complexity. The objects are added to resist against *segmentation-attack*, i.e., the segmentation of the hands cannot be easily possible from a noisy background using some filters. The RGB color ($\tau_{RGB}$) and opacity ($\tau_{opacity}$) of each shape are arbitrarily selected in *step* 3. Now, $I_B$ is undergone with different distortions. A candidate distortion (δ) to perform a specific operation (d) on an image, I, is denoted by $\delta_d(I)$. A particular sequence of successive deformations defines a composite distortion, where, the innermost $\delta_{d_1}$ and outermost $\delta_{d_n}$ denote the first and last distortion, respectively.

$$\delta_{composite} \leftarrow \delta_{d_n}\left(\delta_{d_{n-1}}(\dots\delta_{d_1}(I)\dots)\right) \tag{8}$$

The geometric transformations (translation, rotation, and scaling), noise, intensity correction, morphing, and blending are considered for image deformations in HCA. The shapes with various parametric values are overlaid on $I_B$ at arbitrary positions. The morphological opening is applied using a disk with a variable radius (5-10 units), and the salt and pepper noise with a varying density (0.02-0.05) is added. A cluttered background ($\bar{I}_B$) obtained in *step* 4 is used for further processing, and $\bar{I}_B$ is stored as an image $I_{BG}$ (*step* 5).

#### 3.1.2 Genuine and Fake Hand Image Preprocessing

The dataset $D_G$ is enhanced with 500 individual classes, and every class contains two $I_G$ ($n_G$=2) images. A class-label ($t_{I_G}$) is chosen randomly (*step* 6), and two $I_G$ are selected from that class. Likewise, $D_F$ contains 400 unique fake images and each image is indexed singly with a class-label $u_{I_F}$ (*step* 12). According to specification, $n_F$ number ($5\leq n_F \leq 7$) of random $I_F$ is selected. The $n_F$ is varied for additional randomness to minimize $p_{\mathcal{H}}(Q)$. The dimension of every $I_G$ is variable, and it is limited to a maximum of 150×150 pixels. Variation in dimension mainly, scaling of images, is an effective affine transformation method against pixel-by-pixel attacks. The hand images are rotated with a random angle (±θ) in either direction. The actual hand area in $I_G$ is segmented from background according to intensity variation at the boundary. For this intent, the region of interest (ROI) is selected, and the remnant background is neglected. Any background pixel that exists outside ROI is set to the same intensity as that pixel of $\bar{I}_B$. After ROI selection, $I_G$ has undergone different distortions that are also applied to each $I_F$. Three transformation methods, namely, size ($\delta_{size}$), rotation ($\delta_{rotation}$), and translation ($\delta_{translation}$), are defined through a composite distortion that is applied to $I_G$ (*step* 8) and $I_F$ (*step* 13). The composite distortion is defined as

$$\bar{I}_S^{(i)} \leftarrow \delta_{translation}\left(\delta_{rotation}\left(\delta_{size}\left(I^{i}_{q,S}\right)\right)\right) \tag{9}$$

where, $I^{i}_{q,S}$ represents the $i^{th}$ image, either real (i.e., two $I_G$ of the same hand of a person) or fake (any type of left/right $I_F$ sample) hand S whose label-index is $q$, i.e., $I_S$=$I_G$ or $I_S$=$I_F$; and $\bar{I}_S$=$\bar{I}_G$ or $\bar{I}_S$=$\bar{I}_F$. The space between any two $\bar{I}_S$ is maintained to evade overlapping during translation. It is also ensured that the location of any $\bar{I}_S$ should not exceed the dimension of $I_B$ during mapping for exact positioning. It maintains the minimum spacing condition regarding pixels between any two images and a boundary area. The $S_{HC}$ is logically divided into 3×3 blocks, and each block is assigned with a numeric label vertically, according to column-major order. The distorted real (*step* 9) and fake (*step* 14) hand images are translated and superimposed on the blocks, one image at any single block as the foreground on $\bar{I}_B$. The operations above are followed for each hand image.

#### 3.1.3 Blending and Correction

The α-blending is applied with a variable opacity to generate $I_{HCAPTCHA}$ (*step* 16). The value of α has been varied from 0.1 to 0.8. In most of the cases, α=0.25 to 0.50 produce a good result. It is observed that α with a higher value creates visualization error in some cases. Thus, α=0.25 is chosen. Finally, γ- correction is applied to maintain the luminance of $I_{HCAPTCHA}$ (*step* 17). Luminance variation is a good factor for human recognizability $p_{\mathcal{H}}(Q)$. After several trial tests, the value of γ has been selected and varied from 1.5 to 2.5. It is observed that γ=2.5 can produce acceptable brighter images for most of the darker background. For an $I_{HCAPTCHA}$ with adequate brightness, γ=1.5 is applied. Lastly, a challenging $I_{HCAPTCHA}$ is generated (*step* 18). The time complexity of *Algorithm-1* is **O**($X\times Y$), and it depends on the dimension of $I_{HCAPTCHA}$.

#### 3.1.4 HandCAPTCHA Verification Method

When solving $I_{HCAPTCHA}$, a user has to determine the positions of two $I_G$ of the same person. It can be solved by looking up those two semantically similar hands by specifying its locations (according to the assigned labels in 3×3 blocks) in a given text field of a graphical user interface (GUI) based application containing $I_{HCAPTCHA}$. A text input field is used (like a text-CAPTCHA solving scenario) as an interface to collect the solution from a claimant for simplicity. In the click-based method, similar hands can be located irrespective of the block-indexing method. So, any random indexing method can be used. The valid clickable area regarding pixels on each $I_{HCAPTCHA}$ is random (explained in Sec 3.2.1).

### 3.2 Security Analysis of HandCAPTCHA

The locations of real hands in the grid and internal pixels of the ROI are computationally independent and dynamic. It connotes that each $I_{HCAPTCHA}$ is independent of others to satisfy (4). Thus, a guessing attack is also independent.

**Object segmentation**: According to CaRP (*Captcha as gRaphical Passwords*) (Zhu et al., 2014), the complexity (**O**) of object segmentation (Ś) from an image is exponentially dependent on the number of objects (N) and polynomially dependent on the dimension (D) of N objects. The complexity can be defined as **O**(Ś)=$ß^N$.P(D), where the parameter ß>1, and P(D) is a polynomial function. In HCA, the utmost nine hands can be included and segmented (ignoring the shape artifacts used for segmentation-resistant). So, the complexity is **O**($Ś_{HCAPTCHA}$)= $ß^9$.P(D). In faceDCAPTCHA, at most, six faces can be segmented. Thus, the complexity is **O**($Ś_{faceDCAPTCHA}$)=$ß^6$.P(D). It can be inferred that HCA is $ß^9$.P(D)/$ß^6$.P(D)=$ß^3$ times harder than faceDCAPTCHA. Hand segmentation from an $I_{HCAPTCHA}$ can be interpreted as computationally expensive. However, there is an inevitable trade-off between the number and dimension of the objects used in the algorithm. It is tested that random addition of different shaped objects and noise toughen to segment the contours of real hands using a well-known filter, such as a median filter. These shapes create difficulties for edge detectors to extract contour for bot attacks and reduces $p_{\mathcal{M}}(Q)$.

**False accept rate (FAR):** As an evaluation metric of HCA, the false accept rate ($FAR_{HC}$) represents the probability that a bot can solve an $I_{HCAPTCHA}$, and the false reject rate denotes the probability that a human fails to answer it. A theoretical bound of permissible 1.5% FAR is defined (Osadchy et al., 2017). It depends on the number of possible answers $n_T$ (here, $n_T$ is a total number of hands in $I_{HCAPTCHA}$), and the number of a successful solution of $q$ challenges by a bot is FAR= $(1/n_T)^q$. To maintain the trade-off between security and usability, $n_T$=8 and q=2 have been tested that results in 1.5625% FAR. In HCA, at most $n_T$=9 has been considered with q=2. $FAR_{HC}$ of our method is 1.2345%, which lies within the limit of 1.5%. Thus, HCA achieves better $FAR_{HC}$ and more challenging object segmentation complexity over the existing state-of-the-art IRC.

### 3.2.1 *Automated Online Guessing Attacks*

Mainly, the trial method is used to undermine the strength of underlying HCA in an automated online guessing attack. Guessing the clickable pixels in one $I_{HCAPTCHA}$ is independent of another as internal object points are computationally independent. In this attack, the assumption is that an $I_{HCAPTCHA}$ is used for learning about valid inner pixels, which is used in testing by random guess for click-attack. The hardness is to decide whether a learning $I_{HCAPTCHA}$ is semantically similar to the testable one to a more significant extent to provide a correct solution. The success of this attack model depends on the following:

i) ***Assumption-1:*** Database $D_G$ is accessible to the attacker. The probability of guessing the same hand-type either left or right of a subject from the total number of classes ($N_G$=500) is $p_G = 1/(2\times N_G) = 1/1000 = 1\times10^{-3}$

ii) ***Assumption-2:*** The dimensions of genuine hands in the learned $I_{HCAPTCHA}$ are more significant or the same in the testing $I_{HCAPTCHA}$ so that an attacker has full knowledge about the entire set of valid pixels of interest. It is assumed that in the learning $I_{HCAPTCHA}$, the dimensions of $\bar{I}_G$ are possibly maximum with 150×150 pixels. Alternatively, in testing $I_{HCAPTCHA}$, aspects of $\bar{I}_G$ are minimum 100×100 pixels, which can maximize the success of a machine learning technique. The maximum difference between the dimensions is 51 pixels, including the end pixels. The probability of similarity between the dimensions ($\delta_{size}$) of two real hands is $p_S = (1/\delta_{size})^2 = (1/51\times1/51)^2 = 1.48\times10^{-7}$

iii) ***Assumption-3:*** During testing, genuine hands are overlaid in the same grid indices as in the learned $I_{HCAPTCHA}$. The probability is $p_L = 1/$*permutation* (9, 2) = 1/72 = 0.01388

Thus, the total probability is $p_T = p_G\times p_S\times p_L = 2.054\times10^{-12}$

Now, if it is pessimistically anticipated that only 10% of pixels are adequate using a machine learning method for *Assumption-2*, then, $p_S$= $1.48\times10^{-3}$, and $p_T = 2.054\times10^{-8}$. According to assumptions, the search space of pixels is $10^3\times(51\times51)^2\times72$. Consider a high-speed system which requires only 1ns to test a pixel, the searching time for the entire pixel-set is about 8.1 minutes (Note*:* it takes 2 minutes for an internet of things (IoT) device to be attacked (ISTR, 2017), which is considerably larger than any real CAPTCHA verification session expiry. It implies the aptness of (4) to thwart $p_{\mathcal{M}}(Q)$. Most of the available works on IRC reported that the time should be shorter than the 60s. In our two-fold system, the total time required is lesser than the 16s. (Sec. 6.3.3). Hence, if the response time is more than the 30s, it can be decided that the claimant is a botnet. Thus, after a time-limit threshold, a new session may appear again for verification.

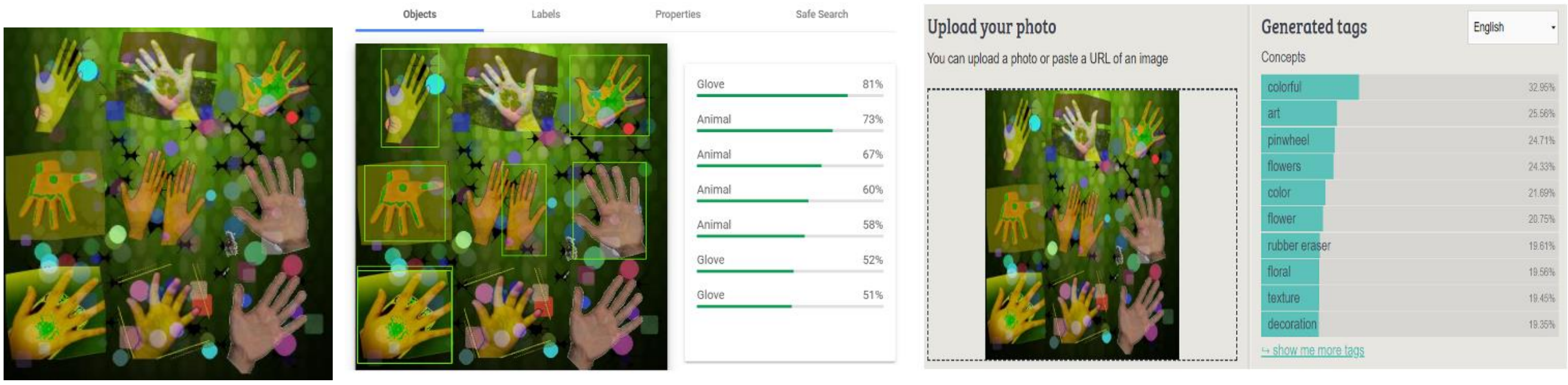

(a) Sample HandCAPTCHA (b) Hand detection with Google Cloud Vision API (c) Imagga Tagging API

**Fig:4.** Sample outcome using deep learning/ machine learning API for automatic detection and tagging in HandCAPTCHA.

### 3.2.2 Deep learning attack

Recent attack schemes on IRC using deep learning techniques successfully achieve satisfactory accuracy (Sivakorn et al., 2016). Alqahtani and Alsulaiman (2020) have proposed an attack method using Imagga automated tagging tool (Imagga, 2019). It uses the WordNet-based Semantic Similarity API to compute the semantic similarity between the tags of sample and candidate images to break a CAPTCHA. In their attack scenario, an IRC challenge consists of different types of objects for tagging, which is useful for a deep learning method to identify the query objects, such as in reCAPTCHA. However, in HandCAPTCHA, only one type of object (i.e., hand) is used; thus, tagging specific object class may not be an effective solution (Fig.4.c.). Instead, our system requires object detection (i.e., real hand) within a specific region. In this case, the distorted real hand detection within a given bounding box can be sufficient to break the system.

We have tested with a few HandCAPTCHA samples to observe the performance of automated object detection using the Google Cloud's Vision API, shown in Fig.4.b. The accuracy of such automated detection is encouraging to be explored for further research. In this example, it reduces the search space of hand detection from 9 objects to 6 and thereby increases the probability of attack (e.g., 1/72 to 1/30) using such a tool. However, in this example, the automated tool fails to detect another real hand due to algorithmic distortions. Therefore, the solution is ineffective in breaking the $I_{HCAPTCHA}$ using an intelligent tool. The probability of attack can be minimized with sound distortions such that object detection and segmentation should not be easily possible. This is a significant vulnerability of any IRC model. However, to guard against such an attack, human authentication (Level 2) can be a beneficial solution. Also, a time-constraint can be imposed to limit such simulated attempts for an attack.

### 3.2.3 Relay Attacks

Relay-attacks, also known as 'sweatshop attacks,' are very difficult to resist in CDAs. Crowdsourcing platforms like Amazon Mechanical Turk offers incentives to solve a CAPTCHA. Skillful conniving humans are being paid for solving thousands of CAPTCHAs at a low cost, such as $1 (Zhu et al., 2014). Any standard protection method against relay attacks can be applied to HCA, such as mentioned in DeepCAPTCHA, which is based on relying on the client's actual IP address with browsing information and timing information (e.g., time-limit threshold for a new session). However, there is still a vulnerability to protect relay-attack even it offers acceptable $FAR_{HC}$. Thus, the PAD and FBV at Level-2 offer an optimistic solution to resist this attack. In a strict time-restricted environment, a licit user has to pass both phases to defend the relay-attack.

### 3.2.4 Probability Estimation of Random Guessing Attacks

Probability estimation of random guess attacks is an essential factor for CDAs. Here, the rectangular and circular methods are described to calculate the probabilities. The size of $I_{HCAPTCHA}$ with X×Y pixels is denoted by $S_{HC}$, and $S_G$ is the size of every real hand with m×n pixels. The probability of choosing randomly any one of the two $I_G$ by the user is

$$p_1 = 2\times S_G/S_{HC} = 2\times(m\times n)/(X\times Y) \quad (10)$$

The set of remaining pixels is determined as

$$S_U = S_{HC} - S_G = (X\times Y - m\times n) \quad (11)$$

The probability of choosing other genuine image is

$$p_2 = S_G/S_U = (m\times n)/(X\times Y - m\times n) \quad (12)$$

The estimated total probability by random guess is $p_S = p_1\times p_2$. Now, altering the $S_G$ for a fixed $S_{HC}$, a range of probabilities can be estimated. The probability of an attack will increase if $S_G$ increases. (see Fig.6, in Sec. 6.1.3). Next, a circular method is followed as a tolerance concerning which a correct click of the user can be considered (Datta et al., 2005). A responder can select around the center of a right hand, and a circular area is regarded as tolerance for selection. Radius (*r*) is varied to estimate the guessing probability. Let $p_3$ be the probability of selecting anyone between the two $\bar{I}_G$

$$p_3 = 2\pi r^2/(X\times Y) \quad (13)$$

According to (11), for the other real hand, the probability is

$$p_4 = \pi r^2/(X\times Y - \pi r^2) \quad (14)$$

The total probability is computed as $p_R = p_3\times p_4$. The $r$ is varied to estimate $p_R$. Theoretically, the relationship between a rectangular and circular area is $r = \sqrt{(m\times n)/\pi}$.

## 4. Image quality metrics ($IQ_m$)

The $IQ_m$ is vital for the quality assessment of generic as well as biometric images. A set of 14 metrics, denoted with $IQ_m^{14}$ are computed in this work. In (Galbally et al., 2014), $IQ_m^{25}$ are tested, and 9 of those are computed here in the same manner, indexed with (a)-(i) in Table 5. We follow the "quality-difference" hypothesis by Galbally et al. (2014) to compute quality metrics such as SSIM, which requires only an input image to extract quality features and can be integrated with the actual feature extraction module. The edge strength similarity metric (Zhang et al., 2013) and wavelet-based $IQ_m$ (Reenu et al., 2013) is indexed with (j) and (k), respectively. The remaining three metrics are proposed here. These are based on the gradient, entropy, and contour profiles, and are indexed with (l)-(n) in Table 5, respectively. These $IQ_m^{14}$ are defined regarding the real hand ($I_G$), and its distorted image ($\hat{I}_G$). Both of the images have the same dimension of X×Y pixels. A low-pass Gaussian kernel (σ=0.5, and size 3×3) is used to degrade $I_G$ that produces $\hat{I}_G$. The quality difference between $I_G$ and $\hat{I}_G$ is computed using a defined metric as $\Delta IQ_m(I_G, I_{\hat{G}}) = IQ_m(I_G) - IQ_m(I_{\hat{G}})$. The proposed $IQ_m$ (minimum disparity implies the highest similarity regarding the metric), denoted by (l)-(n) are defined next. The metrics are listed in Table 5.

**Table 5: Image Quality Metrics Specification ( $IQ_m^{14}$)**

| Property | Quality Metrics |
|---|---|
| Pixel Difference | a) Mean Square Error (MSE), b) Peak Signal to Noise Ratio (PSNR), c) Average Difference (AD), d) Normalized Absolute Error (NAE), e) Structural Content (SC) |
| Correlation | f) Normalized Cross-Correlation (NCC) |
| Structural Similarity | g) Structural Similarity Index Measure (SSIM) |
| Edge | h) Total Corner (*Harris* detector) Difference (TCD), i) Total Edge (*Sobel* operator) Difference (TED) |
| Edge-Strength Similarity | j) Edge-Strength Similarity-Based Image Quality Metric (ESSIM) finds the similarity between the edge-strength maps, representing the anisotropic regularity and irregularity of the edge |
| Wavelet | k) WAvelet based SHarp features (WASH) is based on the sharpness and zero-crossings |
| Gradient | l) Total Gradient Difference (TGD) |
| Entropy | m) Total Entropy Difference (TEnD) |
| Finger Profiles | n)Total Left-side Finger Profile Difference (TFPD) |

(l) *Total Entropy Difference* (TEnD): it merely computes the entropy (H) difference between $I_G$ and $\hat{I}_G$ images.

$$TEnD\left(I_G, \hat{I}_G\right) = \frac{|H(I_G) - H(\hat{I}_G)|}{\max(H(I_G),\ H(\hat{I}_G))} \tag{15}$$

(m) *Total Gradient Difference* (TGD): the finger normalization (Sec. 5) follows an operation that is akin to computing the gradient image based on which TGD is defined.

$$TGD(I_G, \hat{I}_G) = \frac{1}{MN} \sum_{i=1}^{M} \sum_{j=1}^{N} \sqrt{(\mathbb{G}_x - \widehat{\mathbb{G}}_x)^2 + (\mathbb{G}_y - \widehat{\mathbb{G}}_y)^2} \tag{16}$$

where horizontal gradient $\mathbb{G}_x = 0.5 \times |(I_G(i,j) - I_G(i+1,j))|$, and vertical gradient $\mathbb{G}_y = 0.5 \times |(I_G\ (i,j) - I_G(i,j+1))|$; i=1,….,M; j=1,…., N are defined, and $\Delta\mathbb{G} = (\mathbb{G}_x^2 + \mathbb{G}_y^2)^{0.5}$.

(n) *Total Left-side Finger Profile Difference* (TFPD): the finger shape profiles are suitable as a quality metric. The details of extracting the finger profiles (FP) are described in (Bera et al., 2017).

$$TFPD\left(I_G, \hat{I}_G\right) = \frac{1}{MN} \sum_{i=1}^{M} \sum_{j=1}^{N} \sqrt{(FP_x - FP_y)^2 + (\widehat{FP}_x - \widehat{FP}_y)^2} \tag{17}$$

where $FP_x$ defines the left-side of finger profiles (LSFP) of $I_G$ using $\mathbb{G}_x$, and $FP_y$ defines the LSFP of $I_G$ using $\mathbb{G}_y$. A dataset for PAD has been created, based on which the aforementioned $IQ_m^{14}$ are computed. The details are given in Section 6.2.

## 5. Hand Image Normalization and Feature Selection

Image normalization is crucial for hand prototyping to reduce intra-class variations. The preprocessing and finger segmentation methods follow successive steps. First, the necessary steps are followed to extract binary hand contour in a pose-invariant manner with required rotation using an ellipse fitting method. Next, the finger profiles are extracted for segmenting the fingers based on the grayscale image transformation, subtraction, and *xor* operation. These operations result in the left-side of finger profiles (LSFP) and right-side of finger profiles (RSFP), respectively. A reference line is delineated to remove wrist irregularities. Lastly, the conflation of the matching LSFP and RSFP fragment renders a normalized finger that is separated from all other fingers. This finger isolation method is attractive due to its algorithmic simplicity. The crucial steps of finger biometrics with feature selection are ideated in Fig.5.

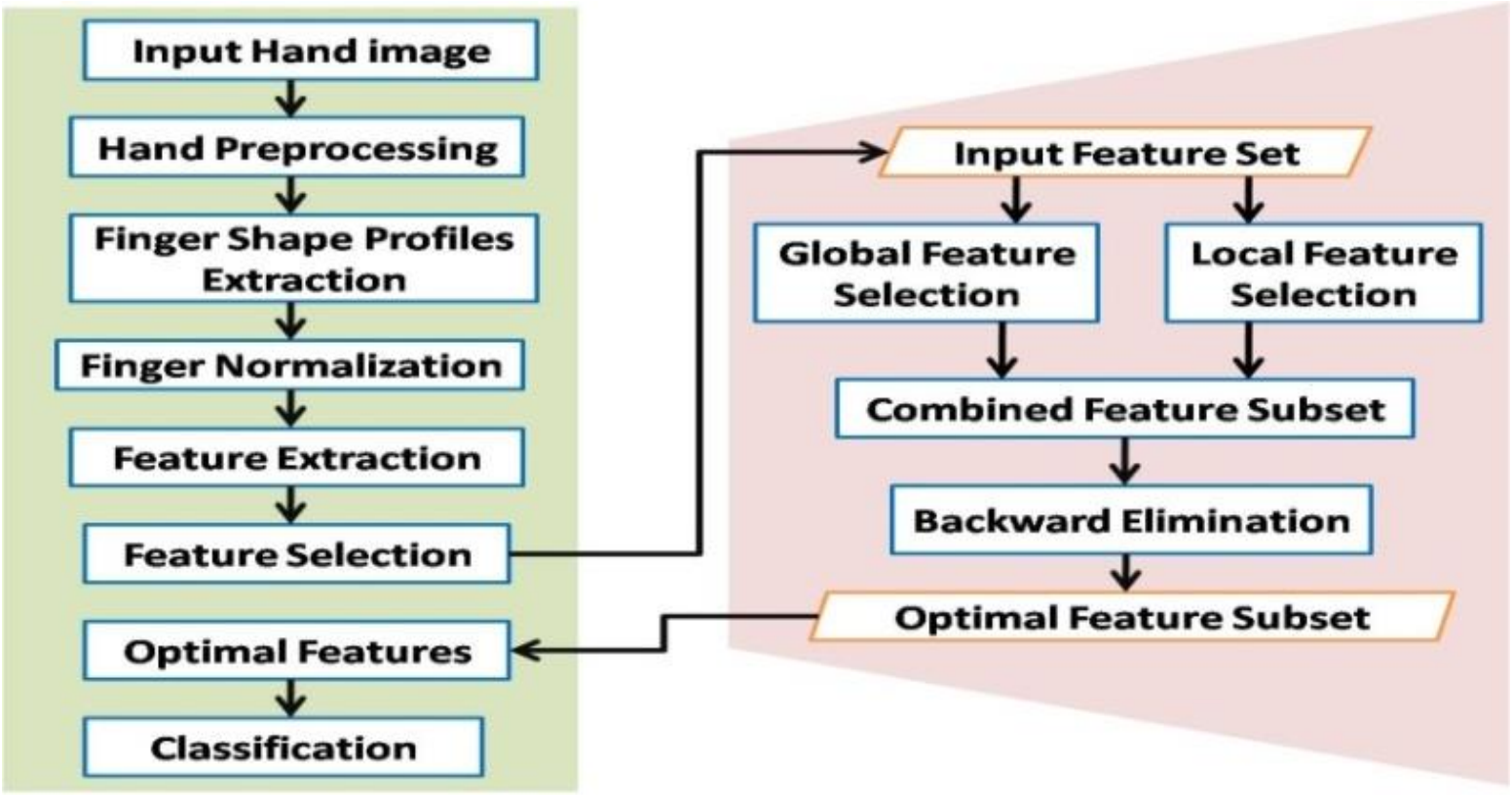

**Fig.5.** Finger biometric system with the M-FoBa feature selection method.

### 5.1 *Finger Geometric Feature Definition*

A few geometric features minimize the computation time and reduce the matching time of FBV. Altogether 26 features from each of the four fingers are computed, defined in Table 6. Mainly, the inaptness of the thumb finger precludes its inclusion for feature extraction (Kang and Wu, 2014). The magnitudes of original features ($f_i$) are in different ranges. It implies that a $f_i$ with high value can dominate another $f_j$ having a low value. It causes other features to be irrelevant during matching. So, the features are normalized to [0, 1] using the *min-max* rule

$$f_{i,j} = \left(f_{i,j} - \min(f_i)\right)/(\max(f_i) - \min(f_i)) \quad (18)$$

where $f_{i,j}$ denotes the $i^{th}$ feature of the $j^{th}$ subject, $max(f_i)$, and $min(f_i)$ denote the maximum and minimum values of the $i^{th}$ feature, respectively, which are determined during training. The $IQ_m$ is also normalized before the experiment using (18).

**Table 6: Finger Geometric Feature Definition**

| | Definition of computed geometric features |
|---|---|
| a) | ***Area:*** the number of boundary pixels of a finger. |
| b) | ***Perimeter:*** the distance around the finger boundary, calculated as the distance between each pair of adjacent contour pixels of a finger. The boundary is considered as a one-pixel wide connected contour. |
| c) | ***Major-axis length*** **and** ***minor-axis length***: the major-axis and minor-axis lengths of an ellipse, which is fitted on each finger. These axes lengths are invariant to translation, rotation, and scaling. |
| d) | ***Equivalent diameter:*** the diameter of a circle with the same area: $(4\times AREA/\pi)^{0.5}$ |
| e) | ***Solidity:*** the ratio of the contour pixels of the finger area and the convex hull. |
| f) | ***Distances from the centroid*** $(x_c,y_c)$ of the finger to ten equal-distant contour pixels is computed as $D_i = [(x_i - x_c)^2 + (y_i - y_c)^2]^{0.5}$ where, $D_i$ is the distance from $i^{th}$ pixel $(x_i,y_i)$. The arc-length parameterization method, within [0-1] range, is followed to locate the pixels which are equal distance apart. |
| g) | ***Widths*** along the length of a finger at ten equal-distant positions. |

### 5.2 *Finger Biometric Feature Selection*

Feature selection is a crucial task to optimize the cardinality of an input feature set $F=\{f_1, f_2, \ldots, f_n\}$ for improving identity matching performance. A feature can be classified mainly as relevant, irrelevant, and redundant (Guyon and Elisseeff, 2003), and specified in Table 7. A feature can be discriminative while assessed alone using a classifier, defined as independently relevant. The relevance at the feature-level is determined independently of other features. A rank $(d)$ is assigned to every $f_i$, according to the order of independent relevance. According to relevance, F can be represented as

$$F = \{\, d_1, d_2, \ldots, d_n | \varphi(f_1) \geq \varphi(f_2) \ldots \geq \varphi(f_n), f_i \in F \,\} \quad (19)$$

**Table 7: Definition of Feature Relevance Characteristics**

| Independently | Conjointly | Decision |
|---|---|---|
| 0 | 0 | Irrelevant |
| 0 | 1 | Conditionally Relevant |
| 1 | 0 | Redundant |
| 1 | 1 | Relevant |

A feature $f_i$ is called conjointly relevant when the inclusion of $f_i$ in the subset S improves the classification accuracy. During forward selection, $f_i$ is included in S when its inclusion with existing S improves the accuracy at least ε, otherwise irrelevant. In backward elimination, if removal of $f_i$ from S decreases the accuracy, then $f_i$ is relevant, else irrelevant. A feature is called redundant if its inclusion or rejection is trivial and hence is rejected. A $f_i$ is selected as relevant if it is both independently and conjointly relevant. A $f_i$ is called conditionally relevant if its relevance exists only in conjunction with other features, otherwise irrelevant. The conditionally relevant feature can play a vital role when only relevant features may not provide satisfactory performance. A limitation of this feature is the dependency on selected features. Also, feature selection may cause an overfitting problem, i.e., the subset may not be feasible during testing on actual data, which degrades accuracy. It may also be viable to obtain multiple solutions of the feature subset optimization problem. The relevance is determined using the *Pearson* correlation coefficient ($\eta$). It computes relevance between two features, $f_i$, and $f_j$, defined as

$$\eta_{i,j} = Cov(f_i, f_j)/\sigma_{f_i}.\sigma_{f_j} \tag{20}$$

where *Cov()* denotes the covariance, and $\sigma$ is the standard deviation. Again, a feature can be defined as either global ($f_g$) or local ($f_l$). Onward, the definitions of the $f_g$ or $f_l$ features are given in the context of four fingers. A feature is called global ($f_g$) when it is relevant and valid for all four fingers. A feature is defined as local ($f_l$) when it is relevant and valid to one or more finger(s), but not for all four fingers. The fingers are arranged as $\psi$ ={*Index, Middle, Ring, Little*}. A $f_g$ is defined as a group of four features

$$f_g = \{f_i^{Index}, f_i^{Middle}, f_i^{Ring}, f_i^{Little}\} \text{ or } f_g = \{\Psi_i\}_{i=1}^{four\ fingers} \tag{21}$$

So, one $f_g$ contains four $f_i$, and its accuracy is given as

$$\forall f_g : \quad \varphi(f_g) \rightarrow \varphi(\{\Psi_i\}_{i=1}^{4-fingers}) \tag{22}$$

where $i$ denotes the same $i^{th}$ attribute for all fingers, otherwise, $f_l$. Contrarily, every $f_l$ is independent of others. The aim is to find a combination of relevant $f_g$ and $f_l$ to attain good accuracy.

### 5.3 *Modified Forward-Backward Feature Selection (M-FoBa) Algorithm*

A combination of forwarding selection and backward elimination algorithms, namely, *Forward-Backward* (FoBa), helps to overcome the inherent drawbacks associated with each method (Zhang, 2011). First, the sequential selection is followed to choose relevant features. Next, backward elimination is applied to the ensued subset rendered in forwarding selection. It results in a subset of selected features. Here, FoBa is applied to obtain $f_g$ and $f_l$ feature subsets as $F_1$ and $F_2$, respectively, based on which an optimal subset $F_{Opt}$ is obtained. Unlike the rank-based FoBa (Bera and Bhattacharjee, 2020), which requires a pre-computation of relevance for all $f_g$ before selection. Here, to avoid pre-calculation, any $f_i$ can be considered randomly as the first selected feature $f_1$. Next, every $f_i$ is tested whether to be included or rejected accordingly concerning marginal accuracy $\varepsilon$=0.5%. As a result, the subset S is obtained, which would be used as input in the removal phase.

The same sequence of $f_i$ is maintained for elimination, as followed during the selection process. However, some different random combinations have been tested, which do not provide remarkable results. *Algorithm-2* is followed to attain $F_{Opt}$ as the simple FoBa may not find an optimal subset for FBV (see Table 12). A simple observation is that the same $f_i$ of a specific finger may not be adequate discriminative compared to the same $f_i$ of another finger. In rank-based FoBa, a $f_g$ may contain irrelevant and/or redundant feature(s) in a group of features. Thus, a combination of both $f_g$ and $f_l$ is followed to minimize the subset cardinality, which is the novelty of M-FoBa. The feature selection method is discussed with details in the next section.

**Algorithm 2:** Modified-FoBa (M-FoBa) feature selection

***Input:*** *Feature set: F*

***Output:*** *Optimal feature subset:* $F_{Opt}$

1. formulate a subset with the global features, the same attribute one per finger: $F_1 \leftarrow$ *FoBa* (F)
2. formulate a subset using a local feature at a time: $F_2 \leftarrow$ *FoBa* (F)
3. sort out the unique features from $F_1$ and $F_2$ as $F_3 \leftarrow$ *unique* ($F_1$, $F_2$)
4. apply backward elimination on every single feature of $F_3$ for optimal subset generation: $F_{Opt} \leftarrow$ *Backward Elimination* ($F_3$)
5. return $F_{Opt}$

## 6. Experimental Results

The experimental methods are three-fold. Firstly, $I_{HCAPTCHA}$ is solved by a human. Next, the PAD is tested, and then, authentication of legitimate claimants is verified with the $F_{Opt}$.

### 6.1 HandCAPTCHA Experiments

#### 6.1.1 HandCAPTCHA Datasets

Three different sets of input images are used in HCA.

i) More than 600 background images are accumulated from the Internet. The dataset $D_B$ includes a variety of high-quality images such as flowers, textures, natural scenes, and other forms of objects.

ii) Dataset $D_G$ contains 500 subjects with two $I_G$ per hand of the BU database. Altogether, $D_G$ includes 1000 left-hand and 1000 right-hand images of 500 persons. These two $I_G$ per hand are named as genuine because these are acquired using the same HP Scanjet scanner. The fundamental characteristics of the $I_{G,}$ such as the dimension and background variations, are similar. Thus, the intra-class entropies of $I_G$ are less and computed next.

iii) A total of 400 fake hand images are collected from the Internet and social sites with significant variations. This dataset contains the images (2-D or 3-D) of celebrity, cartoon, dorsal hand, emoticons, gloved hand, infrared hand, handprint, colorful design on hand, hands of women with different stylish printed design, and others. The generic attributes of each $I_F$ are dissimilar, hence, termed as a fake image. The entropies of $I_F$ are more significant, and inter-class entropy variations are noteworthy.

Based on randomly selected ten subjects, the average difference between the entropies (5) of two $I_G$ of a person is 0.06. The average pair-wise difference between the entropies of ten $I_F$ is 1.5. The average pair-wise dissimilarity between the conditional entropies (6) of ten $I_{HCAPTCHA}$ provided with $I_G$ and $I_F$ is 0.4. The differences are computed in a cyclic order.

#### 6.1.2 Usability Study

Altogether, 562 individuals (participant students: 532; volunteer students: 10; and staff members: 20) have participated in solving $I_{HCAPTCHA}$. The undergraduate (UG) and postgraduate students (PG) of various engineering courses and other disciplines (e.g., Science, Finance, Arts, etc.) have partaken in the verification. The age ranges of UG students are 18-22 years, PG students are within 27 years, and staff officials (who are graduate or postgraduate) are between 25-50 years. Among the participants, 377 (67%) are male, and 185 (33%) are female. The volunteers have communicated to the respondents who are from five different universities and institutions. In a few cases, responses are collected directly thorough a laptop from the persons who do not have any computer system and are not reachable through any communication network during the process. Most of the users have prior experience of CAPTCHA solving at different websites. There is no business or payment for participation in this study. The ethical concerns and privacy of the participants are maintained unbiasedly in our work.

The objective of this research and what would be the role of participants are clearly stated. The instructions about how to solve an $I_{HCAPTCHA}$ through a simple GUI are provided in a text- (appears after login page) as well as verbal- (whenever applicable) mode to the partaker before response collection. Before verification, the candidates have provided their information (e.g., name, age, city, etc.) for our study. The maximum time limit is set to 30s per answer after few initial trials. During this session, one can skip and can solve the next challenge. Lastly, overall feedback (e.g., solution cue, remark) about this experience is collected from them. The accuracy ($A_{HC}$) from the human responses is calculated as

$$\textit{Human accuracy } (A_{HC}) = \textit{total number of correct responses / total number of responses} \quad (23)$$

**Training phase:** Firstly, ten enthusiastic UG ($2^{nd}$ and $3^{rd}$ -year students of Computer Science) volunteers are trained on selecting semantically similar two real hands in an $I_{HCAPTCHA}$, and how to identify correct labels according to positions in a 3×3 logical grid. Besides, five Ph.D. scholars and Professors have provided their responses and suggestions for improvements during the training phase. However, their responses are not included for accuracy assessment. According to their comments received during this period, *Algorithm-1* has been improved. For example, high-intensity variation between the foreground and background should be maintained. In a few challenging cases, horizontal or vertical stripes create visualization difficulties because $\bar{I}_F$ hands are also indistinguishable from $\bar{I}_G$ hands. Thus, the stripes are excluded from *Algorithm-1*.

The experimental configuration of $I_{HCAPTCHA}$ verification described in our earlier work (Bera et al., 2018) has been briefed to maintain its consequences in the present experiments. During training, the volunteers are assigned to solve more than fifty $I_{HCAPTCHA}$, and the accuracy is 94%. For testing, 300 unique $I_{HCAPTCHA}$ are generated, and a total of 600 responses are collected from the 100 students. The testing accuracy is 98.34%. Here, $I_{HCAPTCHA}$ is verified in two phases at different periods with two different testing settings. In the first test, the participants have used their laptops. In the second test, a local server in our laboratory is used for response collection. Altogether, 550 unique $I_{HCAPTCHA}$ are created. The recognizability and perceptual experiences of the users to correctly answer $I_{HCAPTCHA}$ are summarized as the cues in the last row after *Algorithm-1*.

**Testing phase 1:** A total of 124 persons (students: 104 and staff members: 20) have partaken in this test. One hundred students are assigned to solve 500 unique $I_{HCAPTCHA}$. Ten responses from each participant are collected by the volunteers that result in 1000 responses. The remaining four students are instructed to solve 500 distinct $I_{HCAPTCHA}$ individually that counts 2000 answers. Lastly, 400 responses are collected from 20 staff members using the remaining 50 unique $I_{HCAPTCHA}$. Thus, the total valid 3400 responses are collected. Few wrong answers are obtained due to the complex background and dark images. A higher γ-value can abruptly change the brightness and pixel intensities. As a result, 55 incorrect answers are obtained. According to the three unique categories of participants, the average accuracy is 98.37%. The accuracy of each group of users is given in Table 8.

**Table 8: Performance Evaluation of HandCAPTCHA using $\alpha = 0.25, \gamma = 2.5$**

| Phase | $I_{HCAPTCHA}$ | Unique participants | Responses per user | Total valid responses | Total correct responses | Accuracy (%) |
|---|---|---|---|---|---|---|
| Training | 20 | 10 | 10 | 100 | 95 | 95.00 |
| Testing 1 | 550 | 100 | 10 | 1000 | 985 | 98.50 |
| | | 20 | 20 | 400 | 393 | 98.25 |
| | | 4 | 500 | 2000 | 1967 | 98.35 |
| Testing 2 | 550 | 100 | 6 | 596 | 588 | 98.66 |
| | | 432 | 6 | 2581 | 2548 | 98.72 |
| Average | 550 | 100+20+432 | 12 | 3981 | 3926 | 98.61 |

**Testing Phase 2:** The volunteers monitor this test, and they instructed the candidates verbally before the answer collection. Overall, 532 candidates have participated, including those 100 students who participated in the earlier test. A set of 6 unique $I_{HCAPTCHA}$ is randomly assigned to each participant. Thus, a total of 3192 responses are collected. However, few users reported that due to high noise, no similarity between the $\bar{I}_G$ could be found. For example, one of the two $\bar{I}_G$ appears to be clumsy and vague. According to their opinion, 15 responses are discarded to avoid such ambiguity and considered as invalid due to preprocessing fault. This error is 0.47%, which has a negligible impact on assessment. Lastly, a total of 3177 valid responses are counted, including 41 wrong answers. In this case, the average of two distinct types of participants is 98.69%. It implies that γ-correction provides a better result than only with α-blending and adjusting luminance maintains human perception in solving $I_{HCAPTCHA}$.

To avoid any bias (if exists) in solutions, valid responses of unique groups of users are evaluated and mentioned in the last row of Table 8. The groups of participants are not familiar with the $I_{HCAPTCHA}$ solving method previously. In this evaluation, we have considered the correct solutions of 100 students and 20 staff members from Testing 1, and 432 participants from Testing 2. These participants are unique and new to solve $I_{HCAPTCHA}$. Note that 2000 responses from 4 students (Testing 1) and responses from the same 100 users in Testing 2 are neglected to avoid any bias in their responses. The average accuracy is computed from 3981 valid answers and is 98.61%. However, considering the accuracies of all five sub-evaluations in both testing cases, the average accuracy, $A_{HC}$=98.496%. Thus, we have considered the minimum of these two average is $A_{HC} \approx 98.5\%$.

The chances of wrong answers are high when one or two positions are blank. Also, the noise can cause an abrupt change in pixel intensities. It affects a change of color at palm and boundary. Besides, noise can damage and fade real hands while overlaid in $I_{HCAPTCHA}$. Interestingly, if two $\bar{I}_F$ look very similar to $\bar{I}_G$, then the response may be wrong. Thus, clarity should be maintained and must not be degraded during preprocessing. The visual sensitivity and perception of humans recognize the real hands, which are not feasible by an automated script easily (see Fig.4.b).

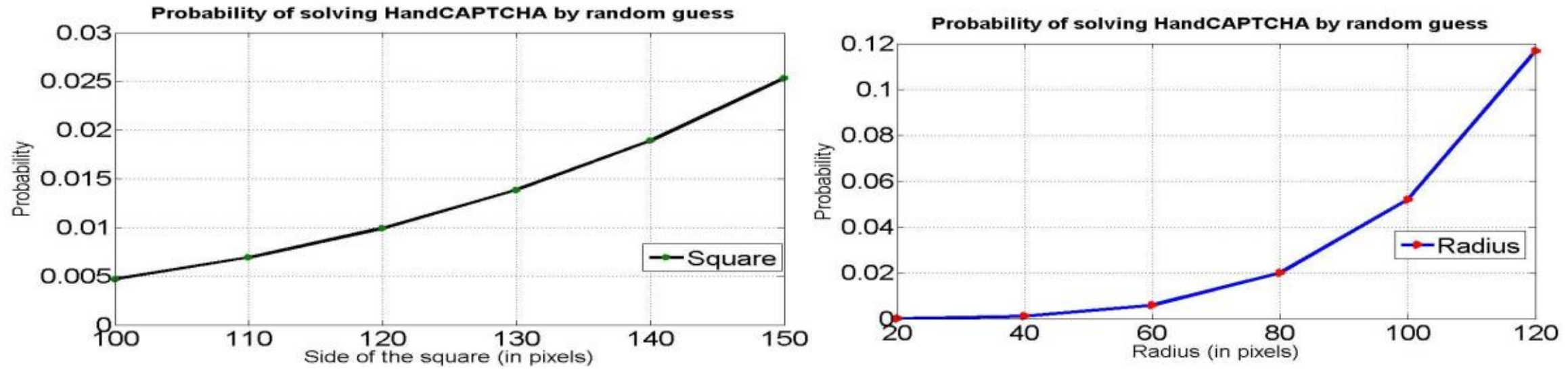

(a) (b)

**Fig. 6.** Probability estimation of solving $I_{HCAPTCHA}$ using different $I_G$ dimensions by random guess-attacks. (a) $I_G$ area: square. (b) $I_G$ area: circle.

#### *6.1.3 Probability Estimation of HandCAPTCHA*

The probabilities of random guess-attack are estimated earlier in Sec.3.2.4. The $S_{HC}$ is set to 460×460 pixels, and the dimension of $S_G$ have been varied from 100×100 to 150×150 pixels. The minimum and maximum probabilities using $p_S$ are $p_{min}$=0.004688, and $p_{max}$=0.025304, respectively. It infers that for a given $S_{HC}$ with a specified higher dimension, and $S_G$ with a lower dimension can render better security regarding probability by random guess endeavor to solve $I_{HCAPTCHA}$. The probability will increase if $S_G$ increases (Fig.6). Though the dimensions of both $\bar{I}_G$ differ during execution, the probability must be within [$p_{min}$, $p_{max}$] for a real situation.

In the circular method, the radius (*r*) has been varied from 20 to 120 units, and $p_R$ has been estimated accordingly. The relation between both methods is compared. For example, to achieve a similar probability (0.00693) as 110×110 pixels, a circle with *r*=62 is required. Based on a random click, without selecting the real hands, 9×8=72 different combinations are possible. Thus, the probability of random selection using a brute-force attack is 1/72= 0.01388. The average $A_{HC}$=98.5% reflects satisfactory user acceptability and usability. The security analysis proves that a bot attack is about 1.23% $FAR_{HC}$, which is lesser than the permissible limit of 1.5% FAR in (Osadchy et al.,2017).

### 6.2 Experiments for Hand Spoofing Attack Detection

Because there is no public hand geometry spoofing database, we have created a fake hand dataset based on an electronic screen display for anti-spoofing simulation. Few subsets of useful metrics are formulated for experiments.

#### *6.2.1 Fake Hand Dataset Creation*

The left-hand images of randomly chosen 255 subjects of the BU database are regarded as the real hand ($I_G$). As the genuine hand dataset comprises with the left-hand images, the fake dataset is created accordingly with those left-hand images. However, the right hand of the participants can also be chosen for both real and fake datasets. The characteristics of the BU database are described in (Dutağaci et al., 2008). Here, the $I_G$ of enrolled persons is displayed on a laptop, based on which high-quality images are captured as the spoofed ($I_{SP}$) samples using a Canon EOS 700D camera. A uniform image acquisition environment (e.g., lighting condition, distance from the screen, etc.) is maintained to capture an $I_{SP}$ from each of $I_G$. The dataset contains 255×3 real, and equivalent 255×3 spoof hand images.

#### *6.2.2 Experimental Results for PAD*

The experiments are regarded as a binary classification to distinguish the real images from fake samples using the *k*-NN and RF classifiers. Two $I_G$ and equivalent two $I_{SP}$ samples per subject are trained, and the remaining one of each $I_G$ and $I_{SP}$ is used for testing. Now, altering the test image one at a time from the three samples, three different test cases are formulated and experimented. Say, for each person, $I_G = \{I_G^1, I_G^2, I_G^3\}$, three training sets are $\{\{I_G^1, I_G^2\}, \{I_G^1, I_G^3\}, \{I_G^2, I_G^3\}\}$ and the testing sets for each training set according to the order are $\{\{I_G^3\}, \{I_G^2\}, \{I_G^1\}\}$, respectively. A similar method is followed for the spoofed samples. The average of three unique cases is computed. The performance is assessed based on the false genuine rate (FGR) and false fake rate (FFR). The FGR denotes the number of $I_{SP}$ samples are classified as genuine. The FFR represents the number of $I_G$ images are classified as fake. The average error rate (AER), a.k.a. half total error rate is AER =0.5×(FGR + FFR).

**Table 9: Anti-Spoofing Error of Each Image Quality Metric**

| Quality | *k*-NN Classifier (%) | | | RF Classifier (%) | | |
|---|---|---|---|---|---|---|
| Metric | FGR | FFR | AER | FGR | FFR | AER |
| a) MSE | 31.63 | 32.81 | 32.22 | 21.56 | 29.67 | 25.62 |
| b) PSNR | 31.63 | 32.81 | 32.22 | 22.35 | 29.80 | 26.07 |
| c) AD | 5.75 | 5.62 | 5.69 | 4.18 | 4.31 | 4.25 |
| d) NAE | 44.44 | 44.96 | 44.70 | 33.20 | 41.17 | 37.19 |
| e) SC | 33.46 | 36.6 | 35.03 | 24.31 | 32.94 | 28.63 |
| f) NCC | 33.59 | 33.33 | 33.46 | 22.74 | 30.33 | 26.53 |
| g) SSIM | 0.91 | 1.17 | 1.04 | 0.52 | 0.98 | 0.75 |
| h) TCD | 44.83 | 56.08 | 50.46 | 53.07 | 32.71 | 42.89 |
| i) TED | 36.99 | 37.25 | 37.12 | 25.88 | 33.20 | 29.54 |
| j) ESSIM | 1.83 | 2.09 | 1.96 | 1.43 | 1.30 | 1.37 |
| k) WASH | 8.89 | 9.41 | 9.15 | 6.40 | 7.58 | 6.99 |
| l) TGD | 12.55 | 13.33 | 12.94 | 8.89 | 10.85 | 9.87 |
| m) TEnD | 45.62 | 44.44 | 45.03 | 31.30 | 42.61 | 36.96 |
| n) TFPD | 32.15 | 33.33 | 32.74 | 23.14 | 31.24 | 27.19 |

During the experiments, the *k*-NN algorithm is used due to its lower time complexity. The RF is a collection of classification trees with higher accuracy of prediction (Breiman, 2001). Every tree is grown independently to predict a decision, and the *bagging* (bootstrap aggregation) technique is used to classify a test vector using the training samples. First, every normalized $IQ_m$ is assessed independently by the classifiers. The classification trees are varied up to 1-100, and the average results are given in Table 9. The RF renders better results than the *k*-NN. The results reflect the relevance of individual quality. For example, the most significant $IQ_m$ is SSIM. Thus, it is chosen at first. Now, each $IQ_m$ is picked one at a time and tested whether AER is reduced or not. Then, merely using the concept of FoBa, few subsets are formulated. For this intent, ESSIM is included with SSIM secondly, which reduces AER than only with SSIM. This process is repeated for all $IQ_m$ to formulate an optimal subset. A few subsets are specified with AER in Table 10. The best subset containing the SSIM, ESSIM, AD, WASH, and NAE renders no classification error, mentioned in the last row of Table 10.

**Table 10: Different Subsets of Five Relevant Quality Features with Errors**

| Feature subset formulation | *k*-NN (%) | | | RF (%) | | |
|---|---|---|---|---|---|---|
| | FGR | FFR | AER | FGR | FFR | AER |
| SSIM, ESSIM, TGD, TEnD, TFPD | 0.39 | 0.39 | 0.39 | 0.26 | 0.13 | 0.20 |
| SSIM, ESSIM, AD, WASH, NCC | 0.13 | 0 | 0.07 | 0.13 | 0.13 | 0.13 |
| SSIM, ESSIM, AD, WASH, MSE | 0 | 0 | 0 | 0.26 | 0.13 | 0.20 |
| SSIM, ESSIM, AD, WASH, PSNR | 0 | 0 | 0 | 0.13 | 0.13 | 0.13 |
| SSIM, ESSIM, AD, WASH, NAE | 0 | 0 | 0 | 0 | 0 | 0 |

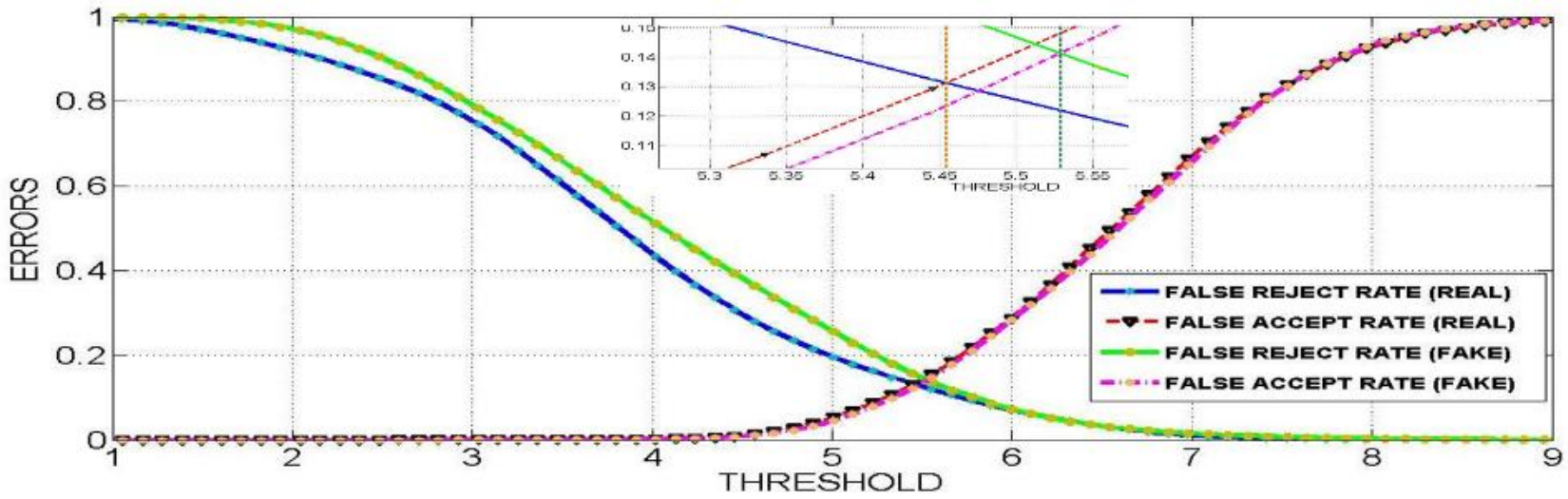

**Fig. 7.** Error estimation regarding the threshold during verification. A zoomed image in the inset implies the thresholds for EERs. For real samples, EER =0.13 at *th*=5.45, and for fake samples, EER =0.14 at *th*=5.53.

The new metrics (TGD, TEnD, and TFPD) do not provide significant performance. These three metrics are tested altogether, and 3.33% AER is obtained using RF. However, these three metrics, along with the SSIM, and ESSIM, provide at most 0.39% and 0.2% AER using *k*-NN and RF, respectively. The standard four $IQ_m$ (SSIM, ESSIM, AD, and WASH) provide 0.13% AER using both classifiers. The experiment for PAD is carried out to discriminate the distributions of selected $IQ_m$ differences between the real and fake samples. Altogether, the $I_G$ (255×2) and $I_{SP}$ (255×2) vectors are trained and tested with 255×1 vectors of $I_G$ and $I_{SP}$. It provides a total of 1020×510=520200 comparisons. These compared differences are considered as the distances between the trained ($q_i$), and test ($\bar{q}_i$) quality vectors. The difference is computed regarding a distance threshold (*th*), defined as

$$th = \sum_{i=1}^{m} \left(\sqrt{(q_i - \bar{q}_i)^2} / \sigma(q_i)\right) \quad (24)$$

where $q_i$ represents the $i^{th}$ quality of trained sample q, and $\bar{q}_i$ means the $i^{th}$ test metric. The standard deviation ($\sigma$) of each metric is calculated over trained samples.

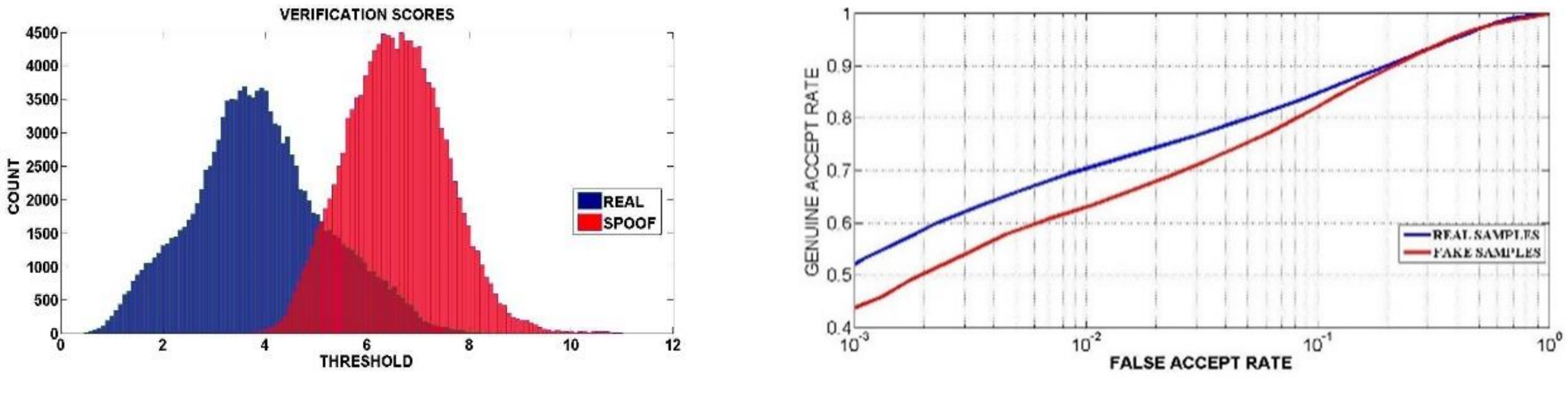

(a) (b)

**Fig. 8.** (a). Verification score distribution of the real (blue) and fake (red) hands for $IQ_m$. (b) The ROC curves are presented using the GAR and FAR.

The errors, namely the false reject rate (FRR) and false accept rate (FAR) of real and fake samples, are computed regarding the threshold. *th* determines the equal error rate (EER) using the errors during verification. A dotted line is emphasized to represent (inset of Fig.7) the intersection point as the EER at a specific threshold. The histogram distributions of the quality differences of real and fake samples are ideated in Fig.8.(a). The receiver operating characteristic (ROC) curves of real and fake samples with the genuine accept rate (GAR) and FAR are shown in Fig.8.(b).

## 6.3 Experimental Analysis of Finger Biometrics

The experiments are conducted on the BU and IITD databases, containing 500 and 200 left-hand subjects, respectively. Three left-hand images per person of the BU dataset are tested to assess the performance. Based on the available three images per hand, two images are preferred for training, and the remaining one image is used for testing at a time as followed for PAD. Three tests are performed by permuting the images for training and testing, and the average accuracies (%) are reported. The experiments are described, preferably based on the BU database as its population is more extensive than IITD. The latter database, IITD (200 subjects), has been tested with the five left-hand images from each subject. Three images per subject are used for training, and the remaining two are selected for testing. Three unique test cases are executed (as mentioned in Sec. 6.2.2) with different combinations of training and testing samples, and the average results are reported.

### *6.3.1 Feature Selection Error during Training*

Feature selection is performed in the training phase using the templates of 100 and 200 subjects, which are chosen randomly from the total population of BU. A *k*-NN classifier evaluates the accuracies. In $F_1$, twelve features per finger, i.e., a total of 12 features×4 fingers=48 features, are selected. Subset $F_2$ contains 46 selected features. The unique features are found out from $F_1$ and $F_2$, and duplicate features are neglected. It results in subset $F_3$, which contains those 68 distinct features. Finally, removal from $F_3$ selects only 35 features in $F_{Opt}$ (see Fig.9.a). Feature overfitting can be handled using a validation test. The M-FoBa is validated by the ten-fold cross-validation error ($E_{CV}$) using the *Pearson* correlation, and *out-of-bag* error ($E_{OOB}$) estimated by the ensemble of classification trees (Breiman, 2001). It computes an average $E_{OOB}$ on the correct estimated prediction used for training. Generally, 90% of trained vectors are randomly chosen, and remnant 10% vectors are used for validation. The average $E_{CV}$ and $E_{OOB}$ are reported in Table 11. The classification trees are varied from 100 to 200. The $E_{OOB}$ decrease while the numbers of the classification tree are more for a fixed number of subjects.

**Table 11: Ten-fold CV-errors and OOB-errors During Training (BU)**

| Subjects | 10-fold CV errors (*k*-NN) | OOB errors (RF) |
|---|---|---|
| 100 | 0.0167 | 0.033 |
| 200 | 0.025 | 0.065 |

**Table 12: Performance Evaluation Using 500 Subjects of the BU**

| Feature subset | $\|F_1\|=48$ | $\|F_2\|=46$ | $\|F_3\|=68$ | $\|F_{Opt}\|=35$ |
|---|---|---|---|---|
| *k*-NN (%) | 96.4 | 96.4 | 95.6 | 98 |

### *6.3.2 Identification*

The subsets ($F_1$, $F_2$, $F_3$, and $F_{Opt}$) during the FS process are evaluated over 500 subjects. The accuracies (Table 12) reflect the suitability of the M-FoBa algorithm. The feature subset formulation is ideated in Fig.9.a. The precision of each feature of $F_{Opt}$ is tested singly to assess the accuracy, shown in Fig.9.b.

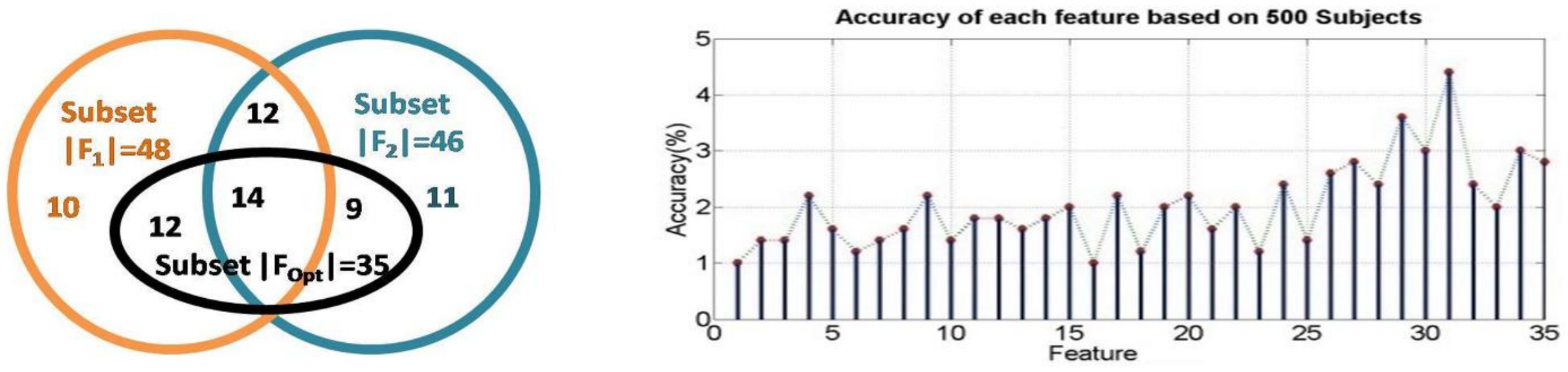

(a) (b)

**Fig. 9.** (a) Feature subset formulation (*Algorithm-2*). (b) Performance evaluation of $F_{Opt}$ by calculating the accuracy (*k*-NN) of each feature independently.

The training and testing cases are evaluated and pictorially ideated in Fig.10. The accuracies of all unique features in $|F_3|=68$ are assessed with the training and testing cases. It is observed that the accuracy of $F_3$ degrades while tested with the remnant features, which are included after the last feature (35th) of $F_{Opt}$ (*see* Fig.10.a). Those 33 unnecessary features are excluded from $F_{Opt}$ using backward elimination. The variations between the accuracies of training and testing datasets are shown to justify the formulation of $F_{Opt}$ in Fig.10.b.

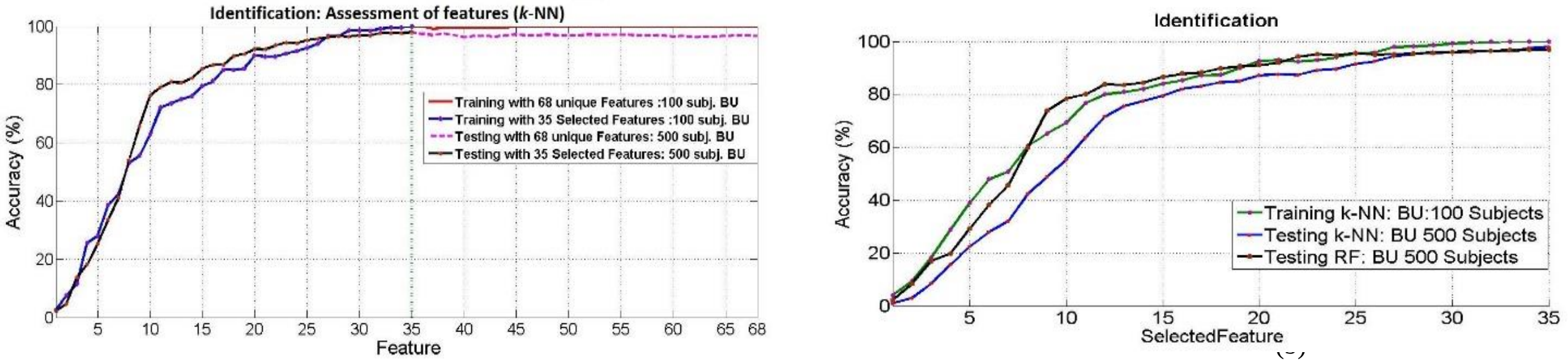

**Fig. 10.** (a) Feature-level assessment on $F_3$ for subset selection (*Algorithm-2*). (b) Identification performance evaluation of $F_{Opt}$ using the *k*-NN and RF.

Next, the subjects are divided randomly into different populations with an increment of 100 subjects per group. In this strategy, several subpopulations are obtained by altering the subjects for each group formation. The average accuracies (%) of identification using the *k*-NN and RF classifiers are given in Table 13.

**Table 13: Performance Evaluation based on $F_{Opt}$ of various populations**

| Database | BU | | | | | IITD | |
|---|---|---|---|---|---|---|---|
| **Subjects** | **100** | **200** | **300** | **400** | **500** | **100** | **200** |
| *k*-NN | 100 | 100 | 99 | 98.7 | 98 | 100 | 99.5 |
| RF | 100 | 100 | 99 | 98.5 | 96.8 | 100 | 98.5 |
| EER | 3.4 | 4 | 4.6 | 5.3 | 6.5 | 5.1 | 5.18 |

The permutation importance ($\mathcal{P}$) rendered by classification trees is suitable for determining the feature relevance (Gregorutti et al., 2017). Here, the learning set $\mathcal{L}_N=\{(F_1,C_1), (F_2,C_2),...,(F_N,C_N)\}$ denotes the feature vector $F_i$ with the associated class label $C_i$, where $F_i=\{f_{i,1}, f_{i,2},..,f_{i,h}\}$, and $h$ is the number of features. $\mathcal{P}$ computes the accuracy of each $F_i$ to predict the class $C_i$ correctly. The predication rule, i.e., the classification scheme is denoted as $\varphi$, and the prediction error regarding the expectation ($\mathbb{E}$) is computed as $\mathcal{P}e = \mathbb{E}[(\varphi(F) - C)]^2$. In this method, a $F_i$ can be considered relevant for predicting the class if breaking the association between the $F_i$ and $C_i$ increases the error $\mathcal{P}e$. For this intent, OOB observations are randomly permuted for each $F_i$. Now, the permuted feature, along with remaining non-permuted features, is employed to predict the response for OOB observations. The difference in prediction accuracy before and after permuting the $F_i$ is averaged over all trees, regarded as a metric for variable importance. The main steps are as follows.

a) Compute the accuracy $\varphi$ of the OOB using a tree $t$.

b) Permute the feature in the OOB observations.

c) Compute the OOB accuracy of $t$ with the feature after permutation once again.

d) Calculate the difference of the predictions between the original (step-b) and recomputed (step-c) OOB accuracy.

e) Repeat the above steps (a-d) for each $t$ and compute the average difference over all the trees as the $\mathcal{P}$ score.

Usually, the $\mathcal{P}$ of a $F_i$ is the score computed as the mean importance over all the classification trees (nT), defined as

$$\mathcal{P}(F_i) = \frac{\sum_{j=1}^{nT}\mathcal{P}\left(F_i^j\right)}{|nT|} = \frac{\sum_{j=1}^{nT}(\varphi(F_i)-C_i)^2}{|nT|} \tag{25}$$

Let, $B_t$ is the OOB sample for a tree t, where $t \in= \{1, \dots, nT\}$. The importance of a $F_i$ in tree $t$ is defined as

$$\mathcal{P}\left(F_i^j\right) = \frac{\sum_{k\in B_t}\varphi(c_k=\bar{c}_k^j)}{|nT|} - \frac{\sum_{k\in B_t}\varphi(c_k=\bar{c}_{k,\pi_i}^j)}{|nT|} \tag{26}$$

where $\bar{c}_k^j$ is the predicted class for the kth observation before permutation, and $\bar{c}_{k,\pi_i}^j$ is the predicted class for the kth observation after permutation of $F_i$ with $f_{k,\pi_j} = \left\{f_{k,1}, \dots, f_{k,i-1},\ f_{\pi_i(k),i},\ f_{k,i+1}, \dots, f_{k,h}\right\}$. The prediction error is computed as

$$\mathcal{P}e(F_i) = \mathbb{E}[(\varphi(F) - C)]^2 - \mathbb{E}[(\varphi(F_i) - C_i)]^2 \tag{27}$$

The OOB importance of every feature is estimated, shown in Fig.11. This is a different method compared to as plotted in Fig.9.b, where simply a $k$-NN classifier estimates the significance at feature-level. This method also implies the aptness of the M-FoBa algorithm for relevant feature selection regarding $\mathcal{P}$. Next, a subject is chosen randomly from the entire population, and the characteristics of related feature vectors are presented in Fig.12. The mean values of normalized features are obtained during training with the entire population, two training sample vectors and one test vector of the chosen subject are plotted to find proximities among the vectors at the feature-level.

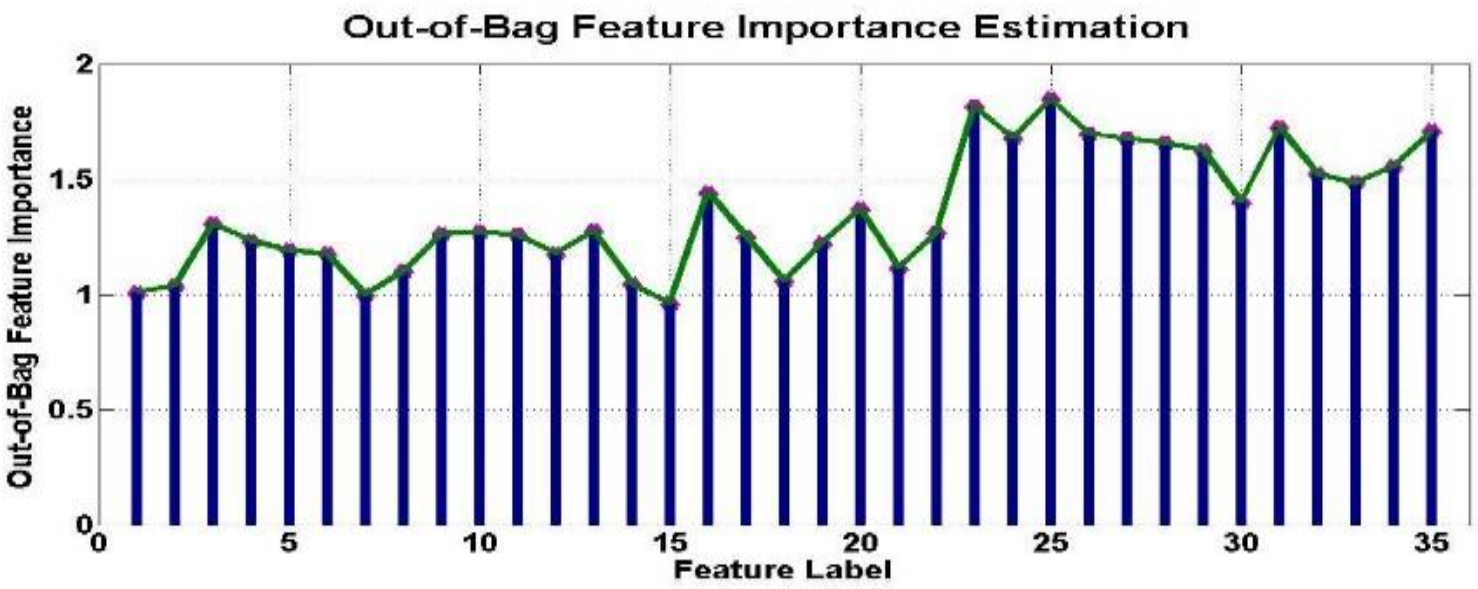

**Fig.11.** Estimation of the OOB-permutation importance using an ensemble. The significance of each feature implies an increase of prediction error if the values of the feature are permuted over the OOB observations. This score is calculated for each tree $t$, and the average value is computed regarding the total ensemble, and lastly, divided by the standard deviation over the whole ensemble. A more significant feature is determined with more substantial value.

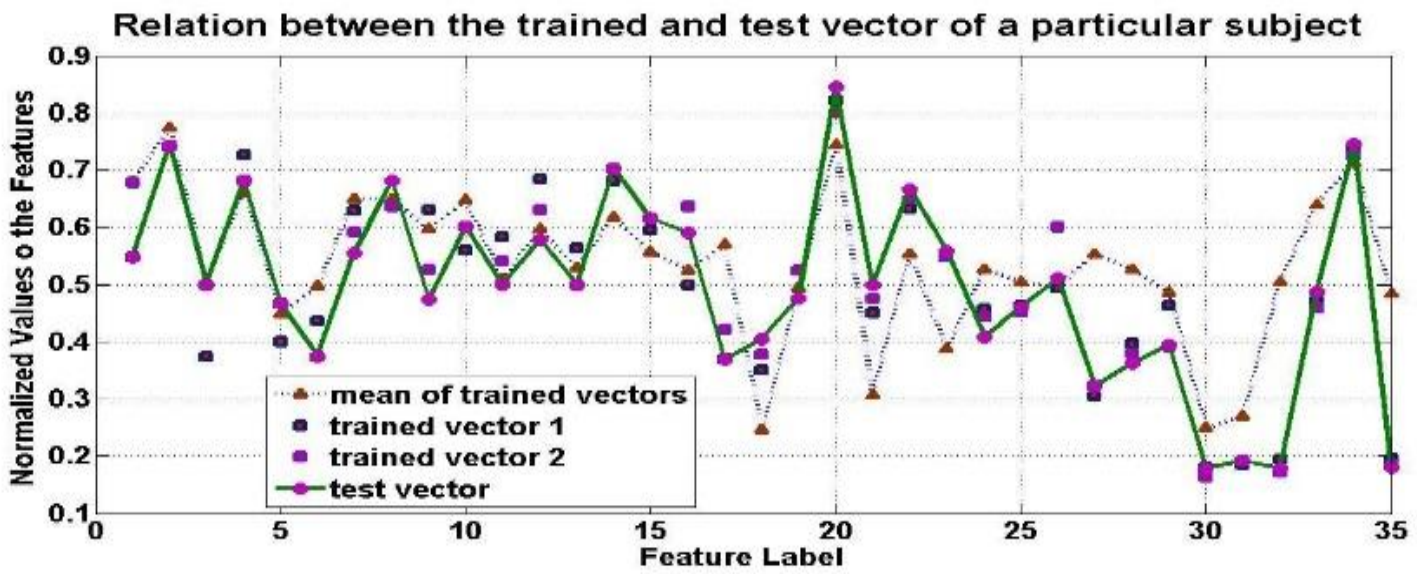

**Fig.12.** Proximity approximation with the trained and testing feature vectors of a randomly chosen from 500 subjects of the BU.

### 6.3.3 *Verification*

The verification accuracy is estimated regarding the EER. A query feature set is matched with the stored templates based on a distance threshold ($th$). The distances between a claimant and registered feature vectors are computed. If the matching similarities exist within $th$, then the person is accepted as a genuine; otherwise, repudiated as an imposter. $th$ is defined as

$$th_{e,g} = \sum_{i=1}^{35}(abs(\, f_{e,i} - b_{g,i})/\sigma(f_i)) \tag{28}$$

where $f_{e,i}$ represents the $i^{th}$ feature of the registered user having serial number $e$ and $b_{g,i}$ mean the $i^{th}$ test feature of a claimant $g$. The standard deviation ($\sigma$) of the $i^{th}$ feature is calculated over the training dataset. The variations of error rates in the ROC curves (Fig.13) between both databases concerning 200 subjects are noteworthy. The EERs for various subpopulations (as formulated during the identification) are mentioned in the last row of Table 13.

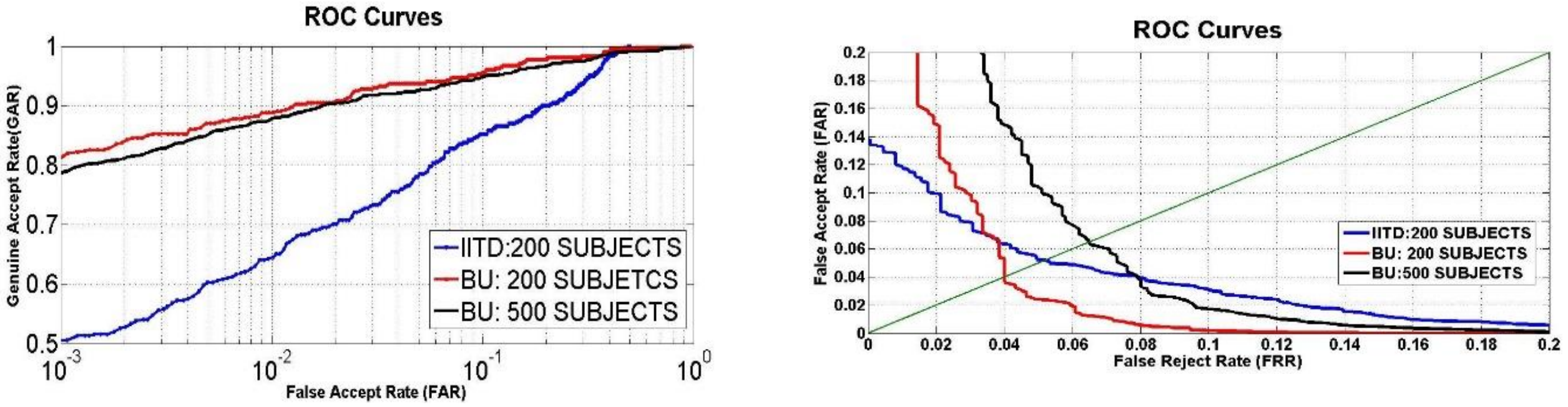

(a) (b)

**Fig.13.** ROC curves. (a) FAR vs. GAR is presented on the logarithmic scale for X-axis. (b) FRR vs. FAR for estimating EER using a diagonal line.

In the next test, verification is performed with a set of disjoint subjects. Some imposter (a.k.a. zero-effort imposter) vectors entirely unknown to the system during this experiment are introduced with the valid subjects. The accuracies for correct verification are given with related comparisons in Table 14. Now, when a new user is enrolled, his/her feature template is stored as genuine, and matching is performed accordingly. In this case, no further training or feature selection is required. Also, it is worth noting that we have selected the feature subset using the samples of the BU dataset, and the same unique features are computed for experiments on the IITD dataset. It implies that no additional training is essential for a new user enrolment or images acquired with a different camera (i.e., dataset variation). However, for a significant modification in the database regarding security, it is essential to update the feature encryption technique (Uludag & Jain, 2004). In this test, the EERs of various cases are more than 6%, comparable to the EERs mentioned earlier in Table 13. In addition to the aforesaid experiments, other test-cases with 500 subjects (BU) using the *k*-NN are observed as follows:

a) Including all the computed (26 features×4 fingers) 104 features, i.e., without feature selection, the accuracy is 87.2%.
b) The accuracy without feature normalization is 93.2%.
c) During elimination, in an intermediate step, the accuracy with 44 features is 97.6%, which retains the same up to 39 features. Those five features are considered as redundant features.

**Table 14: Verification (EER) Performance of Disjoint Subjects**

| **Genuine subjects** | 300 | 300 | 400 |
|---|---|---|---|
| **Genuine comparison** | 300×2 | 300×2 | 400×2 |
| **Imposter subjects** | 200 | 100 | 100 |
| **Imposter test set** | 299+600 | 299+300 | 399+300 |
| **Imposter comparison** | 600×899 | 600×599 | 800×699 |
| **Total comparison** | 600×900 | 600×600 | 800×700 |
| **Equal Error Rate** | 6.3% | 6.1% | 6.2% |

#### *6.3.4 Computation Time*

All the experiments are performed using a system with the Core i3 Processor (2.13 GHz) and 3 GB RAM and coded in MATLAB 2012a. With this system, based on some unique $I_{HCAPTCHA}$, the average computation time is 7.67s. To solve $I_{HCAPTCHA}$ correctly by identifying two real hands, the average time required is about 2s. The average time for $IQ_m$ per image, including the quality feature extraction of five selected metrics (i.e., the best subset for PAD), and classification is 0.3s. The average time required for the preprocessing stage and extraction of selected features for FBV has been estimated as 5.1s. An average of 0.02s per 100 subjects is required to identify a subject. The overall time (<16s) of the proposed work is improved.

#### *6.3.5 Attacks on Hand Biometrics*

Although anti-spoofing is a useful technique for securing a biometric system, there exists vulnerability on any biometric system. Inspired with the "quality-difference" hypothesis by Galbally et al. (2014), we present a fast and straightforward anti-spoofing method on hand biometric by assessing software-based general $IQ_m$. Like their approach, our system requires only a sample biometric image to detect whether it is real or fake. The PAD module can be easily integrated with the feature extractor for spoofing detection. Researchers have already addressed different techniques to tackle this issue.

As a solution, multibiometrics can be a safety measure against attack on a biometric system as the matching scores are obtained from different matching module to make a decision which can reduce the possibility of attack (Uludag & Jain, 2004). Another solution is to limit the number of matching attempts to restrict multiple false matches within a given period. Herein, we limit a time limit (e.g., the 30s) to start a new verification session assuming an intruder attempt to access the system as our system requires 16s for a successful attempt for verification. Another attack can be simulated from the leakage of actual biometric data, which must be secured by an encryption technique. Alternatively, distorted biometric data or feature templates can be a protective measure. If a particular distorted representation of the data template is compromised, a new distortion technique can secure and transform the database. Also, integrating biometric matching and cryptographic techniques to protect the biometric template is another solution (Uludag et al., 2004). However, our proposed approach is not posed to handle such a challenge like actual data leakage, which has already been addressed by the researchers. Instead, in another direction, we tackle automated attacks on a biometric system by leveraging the benefits of CAPTCHA. Thus, protecting the actual biometric data in this proposed system is a significant limitation and is another future direction of research.

### *6.3.6 Performance Comparison*

A quantitative comparison of the performances of the existing IRC is presented in Table 15. The probability evaluation for an arbitrary guessing attack in faceCAPTCHA (Goswami et al., 2012) to localize the genuine faces with 100×100 pixels is 0.688%, and the human accuracy in solving $I_{HCAPTCHA}$ is 98%. Contrarily, the probability of a random guessing attack using HCA for the equal dimension of the image is 0.4688%, and the best accuracy of solving by a person is 98.72%, and an average is 98.5%. Likewise, the probability of guessing attack using face-DCAPTCHA (Goswami et al., 2014) by considering two real faces with a dimension of 400×300 pixels and tolerance size of 80×80 pixels is 0.6%. The probability of the present method using similar experimental constraints is 0.19%. So, the HCA is less vulnerable, with 1.23% $FAR_{HC}$. Notably, γ-correction improves accuracy than only that of α-blending, compared to (Bera et al., 2018). Though the probability is less likely in (Conti et al., 2015), yet, its accuracy and time are not satisfactory. The qualitative advantages of HCA over other IRC methods from the designer aspects can be stated as high randomization of design parameters, less computation time, and less probability of attacks. The chances of human failure in solving $I_{HCAPTCHA}$ depend on the trade-off between blending and illumination correction parameters. The advantages from a user perspective are attributed with quick response time, independent of native language, qualification, age, and ethnicity. These are sustained to maximize $\mathcal{H}(Q)$.

**Table 15: Performance Comparison with Other Works on IRC**

| Ref. | IRC | Accuracy(%) | Probability(%) | Time (s) |
|---|---|---|---|---|
| Torky et al., 2016 | Necklace CAPTCHA | 97.26 | 22.56 | 24 |
| Conti et al., 2015 | CAPTCHaStar | 90.2 | 0.09 | ≈27 |
| Goswami et al., 2012 | faceCAPTCHA | 98 | 0.688 | unspecified |
| Bursztein et al., 2010 | Authorize image turker | 98 | unspecified | 9.8 |
| Bera et al., 2018 | HandCAPTCHA | 98.34 | 0.4688 | unspecified |
| Proposed | HandCAPTCHA | 98.55 | 0.4688 | 9.67 |

The quantitative comparison of the existing hand biometric systems on the BU and IITD datasets are provided in the top-row and bottom-row of Table 16, respectively. The computational complexities and feature spaces in those works are comparatively high due to shape descriptors such as the ICA and SIFT. The proposed method has achieved better EER than the listed state-of-the-art approaches on both databases in Table 16, except (Baena et al., 2013), which was tested on a smaller population with 137 subjects IITD database. The authors have employed the genetic algorithm with hundreds of executions for feature selection, which is computationally expensive for online-based verification regarding computation time. Most of the existing works have targeted to achieve high accuracy, and the computation times have not been reported in their works. It is clear from Table 16 that our approach has achieved the highest accuracies (98% on BU and 99.5% on IITD) than the state-of-the-art methods. Also, the subset $F_{Opt}$ formulated by M-FoBa provides improved accuracies that are competitive over the methods above. Subset cardinality optimization using the M-FoBa algorithm is also remarkable. Lastly, it can be stated that there exists no comparable method for finger biometric anti-spoofing, regardless of the image quality assessment methods in the existing literature.

**Table 16: Performance Comparison with Other Works on Hand Biometrics**

| Ref. | Approach | Subj. | Accuracy (%) | EER (%) |
|---|---|---|---|---|
| Yörük, et al., 2006 | ICA, higher feature space, computationally complex | 458 | 97.81 | GAR: 98.21 |
| El-Sallam et al., 2011 | ICA, higher feature space, score-level fusion is costly | 500 | 97.8 | GAR: 98.5 |
| Bera et al., 2020 | Geometric, Non-optimal subset selection | 500 | 95 | 6.6 |
| Proposed | geometric, optimal subset | 500 | 98 | 6.5 |
| Baena et al., 2013 | Geometric and shape descriptors left | 137 | 98.5 | 4.51 |
| Charfi et al., 2014 | SIFT, computationally complex and costly | 235 | 94 | 5.86 |
| Afifi, 2019 | Deep learning features + Local Binary Pattern | 230 | 94.8 | unspecified |
| Proposed | geometric, optimal subset | 200 | 99.5 | 5.18 |

In summary, the probabilities of $I_{HCAPTCHA}$ and FBV are 0.985 (average) and 0.98 (Table 6: *k*-NN, 500 subjects, BU), respectively. Note that the probability of PAD is one (AER=0, Table 10). The total probability of solving $I_{HCAPTCHA}$ by the human users and biometric authentication with anti-spoofing over the BU dataset is 0.985×1×0.98= 0.9653. Hence, the consolidated accuracy of the proposed verification system is 96.53%.

## 7. Conclusion

An anti-spoofed finger biometric system in conjunction with CAPTCHA verification is presented in this article. The overall system is probabilistically an optimistic solution to offer enhanced security. The HCA is competitive over traditional IRC and advantageous to prevent different malicious attacks. Intuitive advantages of HCA bolster the security and reliability of the FBV. The scalability of HCA implies its applicability to be deployed in the online applications (e.g., social networking) and handheld devices like a smartphone. The $I_{HCAPTCHA}$ solutions from more individuals can be collected in the future.

Additionally, an intelligent algorithm may be devised to assess the performance against the conniving users and different adversarial bots. The FBV module can be tested with other salient features extracted from more population to enhance its robustness in a secure environment. The feature selection algorithm can be improved. The liveness detection using the thermal hand images can be validated for anti-spoofing. In this context, a spoofing database for hand geometry should be created for further research. The FBV of our proposal can be developed with thermal hand images to achieve improved security. This dual-fold verification scheme imparts the security level compared to any single mode reliant on either CAPTCHA or biometrics. It can be devised using any other forms of IRC with other biometric traits (s). However, database security is an important and challenging open issue which should be protected from different attacks. It can be deployed in various commercial applications that require acceptable computation time and FAR of automated attacks.

ACKNOWLEDGMENT

The authors would like to thank the Editor-in-Chief and anonymous Reviewers for their valuable comments. The authors are also thankful to the student volunteers, experts, and participants who have provided their suggestions and responses for this research.

## References

Afifi, M. (2019). 11K Hands: Gender recognition and biometric identification using a large dataset of hand images. Multimedia Tools and Applications. doi.org/10.1007/s11042-019-7424-8

Alqahtani F. H., & Alsulaiman F. A. (2020). Is image-based CAPTCHA secure against attacks based on machine learning? An experimental study. Computers & Security 88, 101635, 1-13

Baena, R. M. L., Elizondo, D., Rubio, E. L., Palomo, E. J., & Watson, T. (2013). Assessment of geometric features for individual identification and verification in biometric hand systems. Expert Syst. with Appl. 40(9), 3580-3594.

Bartuzi E., and Trokielewicz M., (2018). Thermal Features for Presentation Attack Detection in Hand Biometrics. 2018 IEEE 9th International Conference on Biometrics Theory, Applications and Systems (BTAS), pp.1-6.

Belk, M., Fidas, C., Germanakos, P., & Samaras, G. (2015). Do human cognitive differences in information processing affect preference and performance of CAPTCHA? Int. J. Human-Comp. Studies 84, 1-18.

Bera, A., & Bhattacharjee, D. (2020). Human Identification using Selected Features from Finger Geometric Profiles. IEEE Trans. on Systems, Man, and Cybernetics: Systems, 50(3), 747-761.

Bera, A., Bhattacharjee, D., & Nasipuri, M. (2018). Hand biometric verification with hand image based CAPTCHA. 4th Intl.' Doctoral Symposium on Applied Computation and Security Systems, AISC 666, 3-18.

Bera, A., Bhattacharjee, D., & Nasipuri, M. (2017). Finger Contour Profile Based Hand Biometric Recognition. Multimedia Tools and Applications, 76(20), 21451-21479.

Bera, A., Bhattacharjee, D., & Nasipuri, M. (2015). Fusion Based Hand Geometry Recognition using Dempster-Shafer Theory. Intl.' Jrnl. of Pattern Recognition and Artificial Intelligence, 29(5), 1556005/1-24.

Bera, A., Bhattacharjee, D., & Nasipuri, M. (2014). Person Recognition using Alternative Hand Geometry. Intl. Jrnl. of Biometrics, 6(3), 231-247.

Breiman, L. (2001). Random forests. Mach. Learning, 45(1), 5- 32.

Bursztein, E., Bethard, S., Fabry, C., Mitchell, J.C., & Jurafsky, D. (2010). How Good are Humans at Solving CAPTCHAs? A Large Scale Evaluation, IEEE Sympo. on Security and Privacy, 399-413.

Charfi, N., Trichili, H., Alimi, A. M., & Solaiman, B. (2014). Novel hand biometric system using invariant descriptors. IEEE Intl. Conf. on Soft Computing and Pattern Recognition, 261-266.

Chen, H., Valizadegan, H., Jackson, C., Soltysiak, S., & Jain, A. K. (2005). Fake Hands: Spoofing Hand Geometry Systems. Biometric Consortium 2005, Washington DC.

Cheng, Z., Gao, H., Liu, Z., Wu, H., Zi, Y., & Pei, G. (2019). Image-based CAPTCHAs based on neural style transfer. IET Information Security, 13(6), 519-529.

Chingovska, I., dos Anjos, A. R., & Marcel, S. (2014). Biometrics Evaluation under Spoofing Attacks. IEEE Trans. on Information Forensics and Security, 9(12), 2264-2276.

Conti, M., Guarisco, C., & Spolaor, R. (2015). Captchastar! A novel Captcha based on interactive shape discovery. preprint arXiv:1503.00561.

Datta, R., Li, J., & Wang, J. Z. (2009). Exploiting the Human-Machine Gap in Image Recognition for Designing CAPTCHAs. IEEE Trans. on Information Forensics and Security, 4(3), 504-518.

Datta, R., Li, J., & Wang, J. Z. (2005). Imagination: A robust image-based CAPTCHA generation system. Proc. ACM Multimedia, 331-334.

Dutağaci, H., Sankur, B., & Yörük, E. (2008). A comparative analysis of global hand appearance-based person recognition. Journal of Electronic Imaging, 17(1), 011018/1–19.

El-Sallam, A., Sohel, F., & Bennamoun, M. (2011). Robust pose invariant shape-based hand recognition. 6th IEEE Conf. Ind. Electron. Appl. 281-286.

Farmanbar, M., & Toygar, Ö. (2017). Spoof detection on face and palmprint biometrics. Signal, Image and Video Processing, 11(7), 1253-1260.

Galbally, J., Marcel, S., & Fierrez, J. (2014). Image Quality Assessment for Fake Biometric Detection: Application to Iris, Fingerprint, and Face Recognition. IEEE Trans. on Image Processing, 23(2), 710-724.

Gao, H., Cao, F., & Zhang, P. (2016). Annulus: A novel image-based CAPTCHA scheme. IEEE Region 10 Conference (TENCON) Proc. of the International Conference, 464-467.

Gao, H., Lei, L., Zhou, X., Li, J., & Liu, X. (2015). The robustness of face-based CAPTCHAs. IEEE Int. Con. CIT/IUCC/DASC/PICOM, 2248-2255.

Gao, S., Mohamed, M., Saxena, N., & Zhang, C. (2019). Emerging-Image Motion CAPTCHAs: Vulnerabilities of Existing Designs, and Countermeasures. IEEE Trans. Depnd. and Secure Comput., 16(6), 1040-1053.

Goswami, G., Powell, B. M., Vatsa, M., Singh, R., & Noore, A. (2014). FaceDCAPTCHA: Face detection based color image CAPTCHA. Future Generation Computer Systems 31, 59–68.

Goswami, G., Singh, R., Vatsa, M., Powell, B., & Noore, A. (2012). Face Recognition CAPTCHA. 5th IEEE Intl. Conf. on Biometrics: Theory, Applications and System (BTAS), 412-417.

Gregorutti, B., Michel, B., & S-Pierre, P. (2017). Correlation and variable importance in random forests. Statistics and Comput., 27(3), 659-678.

Guyon, I., & Elisseeff, A. (2003). An Introduction to Variable and Feature Selection. Jrnl. of Mach. Learning Research, vol.3, 1157-82.

Google Cloud Vision AI. https://cloud.google.com/vision

Imagga, 2019. Powerful Image Recognition APIs for Automated Categorization & Tagging. https://imagga.com.

ISTR, Symantec. (February, 2019). vol.24. https://www.symantec.com/content/dam/symantec/docs/reports/istr-24-2019-en.pdf

ISTR, Symantec. (March, 2018). vol.23. https://www.symantec.com /content/dam/symantec/docs/reports/istr-23-2018-en.pdf

ISTR, Symantec. (April, 2017). vol.22. https://www.symantec.com /content/dam/symantec/docs/reports/istr-22-2017-en.pdf

Kang, W., & Wu, Q. (2014). Pose-Invariant Hand Shape Recognition Based on Finger Geometry. IEEE Trans. on Sys., Man, and Cybernetics: Systems, 44(11), 1510-1521.

Korshunov, P., & Marcel, S. (2017). Impact of Score Fusion on Voice Biometrics and Presentation Attack Detection in Cross-Database Evaluations. IEEE Journal of Selected Topics in Signal Processing, 11(4), 695-705.

Kumar, A. (2008). Incorporating cohort information for reliable palmprint authentication. 6th Indian conf. on ICVGIP, 583-590.

Kumar, A., & Zhang, D. (2005). Biometric Recognition using Feature Selection and Combination. Proc.:5th Intl.' Conf.' on Audio- and Video-Based Biometric Person Authentication, 813-822.

Madisetty, S., & Desarkar, M. S. (2018). A Neural Network-Based Ensemble Approach for Spam Detection in Twitter. IEEE Trans. on Computational Social Systems, DOI: 10.1109/TCSS.2018.2878852.

Nogueira, R. F., Lotufo, R. A., & Machado, R. C. (2016). Fingerprint Liveness Detection Using Convolutional Neural Networks. IEEE Trans. on Information Forensics and Security, 11(6), 1206-1213.

Osadchy, M., H-Castro, J., Gibson, S. Dunkelman, O., & Pérez-Cabo, D. (2017). No Bot Expects the DeepCAPTCHA! Introducing Immutable Adversarial Examples, With Applications to CAPTCHA Generation. IEEE Trans. on Inform. Forens. and Secrty., 12(11), 2640-2653.

Powell, B. M., Kumar, A., Thapar, J., Goswami, G., Vatsa, M., Singh, R., & Noore, A. (2016). A Multibiometrics-based CAPTCHA for Improved Online Security. IEEE 8th Intl. Conf. on Biometrics Theory, Appl. and Systems.

Raghavendra, R., & Busch, C. (2015). Robust Scheme for Iris Presentation Attack Detection Using Multiscale Binarized Statistical Image Features. IEEE Trans. on Info. Forensics and Security, 10(4), 703-715.

Reenu, M., David, D., Raj, S.S.A., & Nair, M. S. (2013). Wavelet Based Sharp Features (WASH): An Image Quality Assessment Metric Based on HVS. IEEE 2nd Intl. Conf. on Adv. Comput., Netwrk.and Security, 79-83.

Reillo, R.S., Avila, C. S., & Macros, A. G. (2000). Biometric identification through hand geometry measurements. IEEE Trans. on Pattern Analysis and Machine Intelligence, 22(10), 1168–1171.

Sedhai S., & Sun, A. (2017). Semi-Supervised Spam Detection in Twitter Stream. IEEE Trans. on Computational Social Systems, DOI: 10.1109/TCSS. 2017.2773581.

Sharma, S., Dubey, S. R., Singh, S. K., Saxena, R., & Singh, R. K. (2015). Identity verification using shape and geometry of human hands. Expert Systems with Applications, 42, 821–842.

Sivakorn S., Polakis I., & Keromytis, A. D. (2016). I am robot: (deep) learning to break semantic image captchas. In: 2016 IEEE European Symposium on Security and Privacy (EuroSP), pp. 388–403.

Sultana, M., Paul, P. P., & Gavrilova, M. L. (2017). User Recognition from Social Behavior in Computer-Mediated Social Context. IEEE Trans. on Human-Machine Systems, DOI: 10.1109/THMS.2017.2681673.

Tang, M., Gao, H., Zhang, Y., Liu, Y., Zhang, P., & Wang, P. (2018). Research on Deep Learning Techniques in Breaking Text-Based Captchas and Designing Image-Based Captcha. IEEE Trans. on Information Forensics and Security, 13(10), 2522-2537.

Tolosana, R., Barrero, M. G., Busch, C., Ortega-Garcia, J. (2019). Biometric Presentation Attack Detection: Beyond the Visible Spectrum. IEEE Trans. on Info. Forensics and Security, DOI 10.1109/TIFS.2019.2934867.

Torky, M., Meligy, A., & Ibrahim, H. (2016). Securing Online Social Networks against Bad bots based on a Necklace CAPTCHA Approach. IEEE 12th Intl.' Computer Engineering Conference (ICENCO), 158-163.

Uludag U., & Jain, A. K. (2004). Attacks on Biometric Systems: A Case Study in Fingerprints. Proc. SPIE-EI Security, Steganography and Watermarking of Multimedia Contents VI, San Jose, CA, pp. 622–633.

Uludag, U., Pankanti, S., Prabhakar, S., & Jain, A. K. (2004). Biometric cryptosystems: issues and challenges. Proc. of the IEEE (Special Issue on Multimedia Security for Digital Rights Management) 92(6), 948-960.

Uzun, E., Chung, S. P. H., Essa, I., & Lee, W. (2018). rtCaptcha: A Real-Time CAPTCHA Based Liveness Detection System. Network and Distributed Systems Security Symposium, DOI: doi.org/10.14722/ndss.2018.23253.

Yörük, E., Konukoğlu, E., Sankur, B., & Darbon, J. (2006). Shape-Based Hand Recognition. IEEE Trans. on Img. Processing, 15(7), 1803-15.

Zhang, T. (2011). Adaptive Forward-Backward Greedy Algorithm for Learning Sparse Representations. IEEE Trans. Info. Theory, 57(7), 4689-4708.

Zhang, X., Feng, X., Wang, W., & Xue, W. (2013). Edge Strength Similarity for Image Quality Assessment. IEEE Signal Pros. Lett., 20(4), 319-322.

Zhu, B.B., Yan, J., Bao, G., Yang, M., & Xu, N. (2014). Captcha as Graphical Passwords-A New Security Primitive Based on Hard AI Problems. IEEE Trans. on Inform. Forensics and Security, 9(6), 891-904.

Zi, Y., Gao, H., Cheng, Z., & Liu, Y. (2020). An End-to-End Attack on Text CAPTCHAs. IEEE Trans. Info. Forensics and Security, vol.15, 753-766.